%% file: main.tex
\documentclass[journal]{vgtc}                     

\usepackage{microtype}                 
\PassOptionsToPackage{warn}{textcomp}  
\usepackage{textcomp}                  
\usepackage{mathptmx}                  
\usepackage{times}                     
\usepackage{cite}                      
\usepackage{tabu}                      
\usepackage{booktabs}                  
\usepackage[svgnames]{xcolor}
\usepackage[most]{tcolorbox}

\onlineid{0}

\vgtccategory{Research}

\vgtcpapertype{application/design study}

\newcommand{\graph}{\textit{Co-Occurrence Graph}}
\newcommand{\scatterplot}{\textit{Image Segment}}
\newcommand{\steer}{\textit{Interactive Caption}}

\title{Visual Analytics for Efficient Image Exploration and\\ User-Guided Image Captioning}

\author{%
  Yiran Li, Junpeng Wang, Prince Aboagye, Michael Yeh, Yan Zheng, Liang Wang, Wei Zhang, Kwan-Liu Ma
}

\authorfooter{
  \item
  	Yiran Li and Kwan-Liu Ma are with University of California, Davis.
  	E-mail: \{ranli, klma\}@ucdavis.edu
  \item
  	Junpeng Wang, Prince Aboagye, Michael Yeh, Yan Zheng, Liang Wang, and Wei Zhang are with Visa Research.\\
  	E-mail: \{junpenwa, priaboag, miyeh, yazheng, liawang, wzhan\}@visa.com
}

\teaser{
\centering
 \includegraphics[width=\columnwidth]{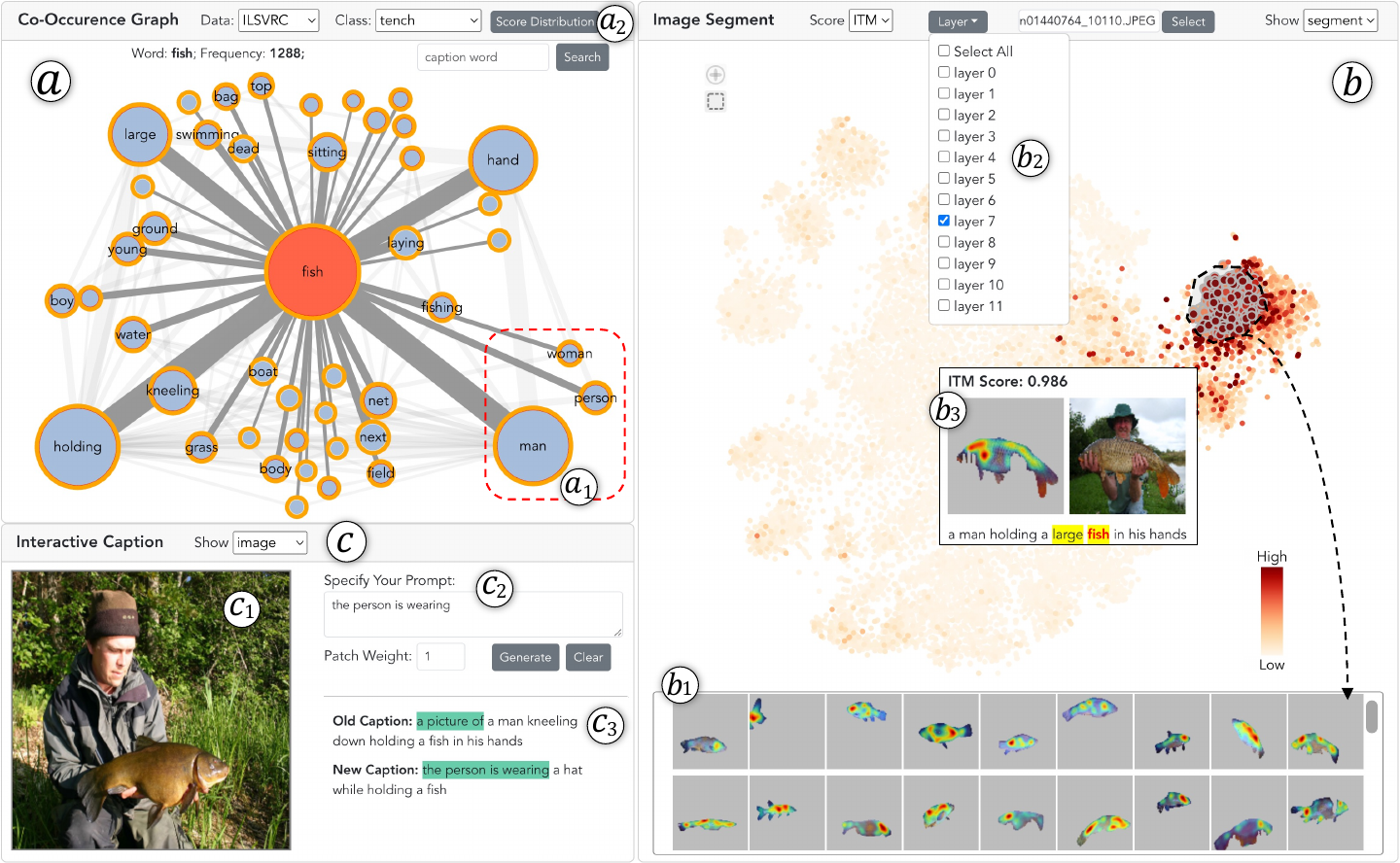}
 \vspace{-0.25in}
 \caption{Our system contains three main views. (a) The \graph{} employs a node-link diagram to present the dominant and co-occurred features of an image dataset. (b) The \scatterplot{} view uses a scatterplot to lay out image segments and color them based on the attention they received. (c) The \steer{} view provides an interface to steer the caption generation process.}
 \label{fig:teaser}
}

\definecolor{ronecolor}{rgb}{0.937, 0.745, 0.173}
\newtcbox{\goalbox}{on line,
  colframe=white,        
  colback=ronecolor,
  coltext=white,        
  boxrule=0.1pt,        
  arc=2pt,              
  boxsep=0.1pt,
  left=2pt,right=2pt,top=1pt,bottom=1pt,
}

\definecolor{rtwocolor}{rgb}{0.518, 0.671, 0.314}
\newtcbox{\taskbox}{on line,
  colframe=white,        
  colback=rtwocolor,
  coltext=white,        
  boxrule=0.1pt,        
  arc=2pt,              
  boxsep=0.1pt,
  left=2pt,right=2pt,top=1pt,bottom=1pt,
}

\input{tex/0abstract}

\begin{document}


\firstsection{Introduction}

\maketitle

\input{tex/1introduction}

\input{tex/2relatedwork}
\input{tex/3background}

\input{tex/4requirements}
\input{tex/5system}
\input{tex/6casestudy}
\input{tex/7discussion}
\input{tex/8conclusion}

\bibliographystyle{abbrv-doi-hyperref}

\bibliography{template}

\clearpage
\appendix 
\input{tex/A_image_exploration}

\input{tex/B_caption_steering}

\input{tex/C_batch_processing}

\input{tex/D_gradcam_evaluation}

\end{document}

%% file: tex/0abstract.tex
\abstract{
Recent advancements in pre-trained large-scale language-image models have ushered in a new era of visual comprehension, offering a significant leap forward.
These breakthroughs have proven particularly instrumental in addressing long-standing challenges that were previously daunting. 
Leveraging these innovative techniques, this paper tackles two well-known issues within the realm of visual analytics: (1) the efficient exploration of large-scale image datasets and identification of potential data biases within them; (2) the evaluation of image captions and steering of their generation process. On the one hand, by visually examining the captions automatically generated from language-image models for an image dataset, we gain deeper insights into the semantic underpinnings of the visual contents, unearthing data biases that may be entrenched within the dataset. On the other hand, by depicting the association between visual contents and textual captions, we expose the weaknesses of pre-trained language-image models in their captioning capability and propose an interactive interface to steer caption generation. The two parts have been coalesced into a coordinated visual analytics system, fostering mutual enrichment of visual and textual elements. We validate the effectiveness of the system with domain practitioners through concrete case studies with large-scale image datasets.
} 

\keywords{Machine learning, visual analytics, data-centric AI, pre-trained language-image models, image captioning}

%% file: tex/1introduction.tex
Recently, the rapid advancements of large-scale machine learning (ML) models have been mirrored by the success of pre-trained language-image models, such as ALBEF~\cite{li2021align}, CLIP~\cite{radford2021learning}, GIT~\cite{wang2022git}, and BLIP~\cite{li2022blip}. These models take a large number of image-text pairs (i.e., images and their captions) as input and learn the correspondence between the two data modalities. Pre-trained language-image models can be used in many downstream tasks, such as zero-short image classification, image captioning, visual question answering, and image retrieval. The profound potential of these models and their wide adoption open promising avenues in the realm of multi-modal learning.

Borrowing the power of these pre-trained language-image models, this work employs visual analytics to approach two common challenges faced by domain practitioners relating to the exploration of image datasets and the quality of generated captions. \textbf{\textit{First}}, when training ML models on an image dataset, domain practitioners often need to obtain an efficient overview of the dataset and be mindful of potential issues rooted in the dataset. For instance, in the LIME paper~\cite{ribeiro2016should}, the authors illustrated how a \texttt{husky dog} image was misclassified as a \texttt{wolf} due to the presence of snow in the image's background. This happened because, in the training set of \texttt{wolf} images, snow was a consistently co-occurring feature. It led to the model developing a bias, assuming any image with snow to be a \texttt{wolf}. Preventing such biased ML models necessitates efficiently exploring large-scale image datasets and recognizing possible data biases before model training, aligning with the rise of data-centric AI~\cite{strickland2022andrew,wang2023visual}. 
\textbf{\textit{Second}}, focusing on the captions generated from image captioning models, there are cases where certain desired image features are not covered. For example, the caption of an image with a \texttt{cat} and a \texttt{dog} may only include descriptions of the \texttt{dog} due to various reasons (e.g., the \texttt{dog} may take more pixels). However, the \texttt{cat} may be of more interest to the users. Recognizing such cases and summarizing their common patterns are vital to understanding the models' captioning behaviors, evaluating the models, and even improving them. However, it is challenging to manually scrutinize individual cases, collect them, and extract their common patterns. 

Due to the widespread nature of these two challenges, several existing studies have proactively attempted to address them. For the first one, DendroMap~\cite{bertucci2022dendromap} hierarchically organizes and presents large-scale image datasets in a tree-map based layout, allowing users to grasp a quick overview of enormous number of images and uncover potential data biases. Focusing on tabular data, CoFact~\cite{kaul2021improving} generates counterfactual examples for data instances with selected features and compares the distributions of the two sets of data to shed light on bias-related data issues. 
Nevertheless, extracting data insights from these works requires considerable effort in data pre-processing and in-depth data exploration.
For the second challenge, various quantitative metrics have been proposed to measure the quality of generated captions. For example, when ground-truth captions are available, BLEU~\cite{papineni2002bleu} and CIDEr~\cite{vedantam2015cider} are frequently utilized to assess the caption quality. However, these metrics are often overly aggregated, providing limited insights into where the weaknesses of a captioning model reside (e.g., what image features were often overlooked by the model?). Furthermore, in cases where ground-truth captions are not available or users wish to generate captions for specific image features, these metrics prove inadequate. 

In this paper, we propose a visual analytics attempt to approach these two challenges. For the first one, we harness the capabilities of pre-trained language-image models, specifically BLIP~\cite{li2022blip}, to generate captions for large-scale image datasets. These captions textualize the image content with descriptive words. We then extract key words out of the captions and build a node-link diagram to demonstrate them and their co-occurrence frequency. The diagram provides an efficient summary of the dataset, guiding detailed exploration of image features. The feature co-occurrence from the diagram further reveals potential data biases. For the second challenge, we decompose images into semantic image segments and reflect their associations with different words from the captions, utilizing the cross-attention of BLIP. By clustering the image segments and depicting their association strengths with different words, we reveal the dominant and less-covered image features. Additionally, we also provide an interface for users to interactively emphasize the overlooked visual features and steer the caption generation process. All above functionalities have been integrated into a coordinated visual analytics system with three main components, as shown in Fig.~\ref{fig:teaser}. We use the system to explore two real-world large-scale image datasets (ImageNet~\cite{imagenet} and COCO~\cite{coco}) and share unique and rich insights that can be gleaned specifically from our system. In summary, our contributions in this paper include:
\begin{enumerate}
    \item An approach to effectively exploring large-scale image datasets and disclosing potential data bias rooted in the datasets.
    \item A visual analytics approach that intuitively presents the association between texts and images, revealing the weaknesses of pre-trained language-image models in visual feature coverage.
    \item A human-in-the-loop solution to steer the existing caption generation process, generating descriptions for desired visual features.
\end{enumerate}

%% file: tex/2relatedwork.tex
\section{Related Work}

Our work is intricately linked to the exploration of large-scale image datasets, the diagnosis and steering of ML models, and the interpretation of Transformers' attention mechanisms. Therefore, we critically examine previous works from these three perspectives.

\textbf{Large-Scale Image Data Exploration.}
Exploring and comprehending large-scale image datasets fall within the purview of \textit{data-centric AI}, which improves ML models by better understanding the training data and enhancing their quality~\cite{strickland2022andrew, wang2023visual}. 
In visual analytics, there has been a considerable amount of works focusing on efficient image exploration. These endeavors can generally be categorized into two groups, exploring images using their performance from ML models or their predominant features.
For the first group, Squares~\cite{ren2016squares} encodes image instances as stacked boxes and analyzes them based on their prediction scores. Blocks~\cite{bilal2017convolutional} groups images by their original and predicted labels, and presents the under-performing groups in a hierarchical confusion matrix. SliceTeller~\cite{zhang2022sliceteller} generates image subsets (slices) based on annotated metadata, and focuses on analyzing under-performing data slices. 
For the second group, EmbeddingProjector~\cite{smilkov2016embedding} presents images through dimensionality reduction algorithms and scatterplot visualizations. Similarly, Karpathy~\cite{karpathytsne} employs tSNE to project a large number of images and uses a grid layout to arrange them to minimize overlaps. DendroMap~\cite{bertucci2022dendromap} adopts an interactive tree-map to facilitate multi-level image explorations and unveil potential data biases within the images.
In contrast to these approaches, our work transforms images into the text domain through image captioning and conducts analyses on the texts to provide insights into the original images. Our solution does not rely on pre-existing performance from ML models or pre-defined image attributes. Also, the modality transformation from images to texts can be largely automated and the text analysis is less labor-intensive compared to existing image exploration approaches.

\textbf{ML Model Diagnosis and Steering.}
\textit{Model diagnosis} is the process of reasoning on unexpected model behaviors and/or uncovering model weaknesses~\cite{yuan2021survey}. For example, Seq2seq-Vis~\cite{strobelt2018s} dissects the execution of a Seq2Seq model into five stages and debugs potential issues at individual stages through customized visualizations.
DQNViz~\cite{wang2018dqnviz} employs a series of chart visualizations and pattern-matching algorithms to visually summarize the behavior patterns of a reinforcement learning agent. 
These patterns are instrumental in unveiling the agent's playing strategy and diagnosing instances where it fails.
AEVis~\cite{cao2020analyzing} analyzes CNNs' weaknesses against adversarial attacks by comparing the models' behavior on benign and adversarial instances.
In this work, we diagnose pre-trained language-image models to reveal their weaknesses in image captioning. By summarizing the coverage of generated captions over image features, we provide a new angle to assess the captioning models.
\textit{Model steering}~\cite{yuan2021survey} empowers users to directly engage with an ML model and control its execution behavior by injecting human knowledge. For example, Yang et al.~\cite{yang2020interactive} proposed ReVision to interactively construct hierarchical clustering trees. In the construction process, users can infuse both public and private knowledge to interactively supervise the tree hierarchy. 
Ming et al.~\cite{ming2019protosteer} used representative prototypes to interpret ML models. Their ProtoSteer system enables users to interactively add, remove, or modify prototypes to enhance interpretability.
Strobelt et al.~\cite{strobelt2021genni} introduced GenNI that employs a human editable constraint graph to better guide text generations. 
More model steering examples also include~\cite{yang2022diagnosing, jia2021towards, chen2021towards}.
In this work, we provide an interface to steer image captioning models. The interface allows users to dynamically emphasize different visual objects within images. Moreover, the promptly generated captions also provide users with immediate feedback to iterate the progressive generation process.

\textbf{Visualization of Transformers' Attentions.}
The attention mechanisms~\cite{bahdanau2014neural}, especially their application in Transformers~\cite{vaswani2017attention}, have achieved great success in the past decade, yielding numerous powerful language~\cite{devlin2019bert, openai2023gpt4, touvron2023llama} and vision~\cite{dosovitskiy2020image, khan2021transformers} models. 
Meanwhile, various techniques have been utilized to visualize attentions. The most popular ones include flow maps~\cite{dong2020interactive, strobelt2018s}, parallel coordinates plots~\cite{vig2019multiscale}, and heatmaps~\cite{jaunet2021visqa, park2019sanvis, aflalo2022vl}. 
These techniques, employed in multiple coordinated views, help to summarize attention patterns and generate insights into their working mechanisms. 
For example, Li et al.~\cite{li2023does} used dimensionality reduction, heatmaps, and scatterplots to effectively summarize attention patterns of vision Transformers (ViTs). These patterns then helped them diagnose different behaviors of the ViTs.
Park et al.~\cite{park2019sanvis} used heatmaps to distinguish attention patterns in a language Transformer and assist the probing of \textit{query} and \textit{key} features within each attention head.
Jaunet et al.~\cite{jaunet2021visqa} explained a visual question-answering Transformer through attention heatmaps, exploring potential biases within it.
Yeh et al.~\cite{yeh2023attentionviz} analyzed different heads' attention behavior by jointly performing dimensionality reduction on the \textit{query} and \textit{key} embeddings.
The interpretation of attentions can also be used to compare different attention heads~\cite{wangDodrioExploringTransformer2021} or different
ML models~\cite{derose2020attention}.
In our research, we interpret the cross-attentions of a pre-trained language-image model and use them to effectively explore one data modality (images) through the other (texts). 
Our interpretation also helps to disclose the weaknesses of pre-trained language-image models.

%% file: tex/3background.tex
\section{Background (BLIP and SAM)}
\label{sec:background}

\begin{figure}[tbh]
 \centering 
\includegraphics[width=\columnwidth]{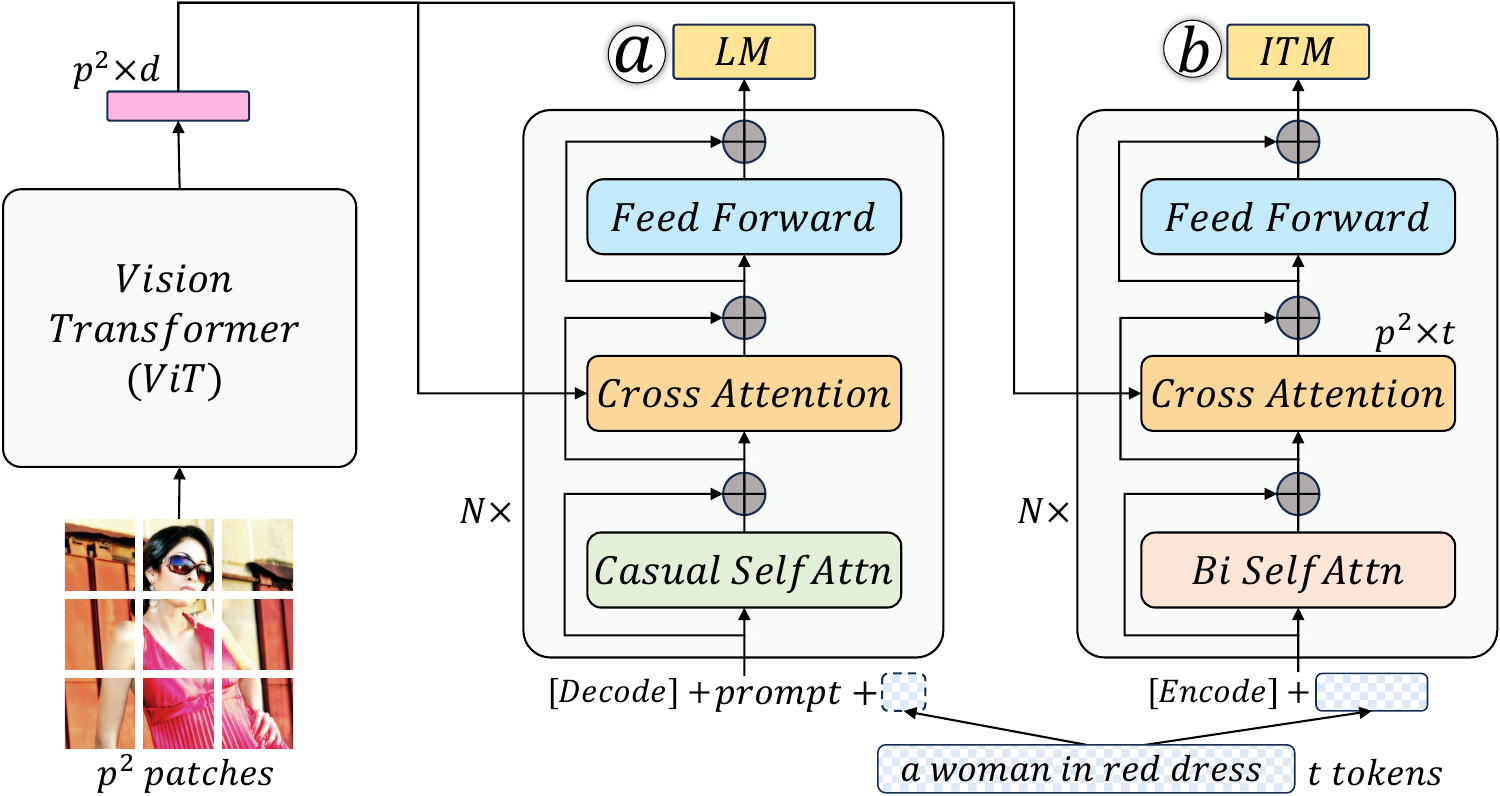}
  \vspace{-0.2in}
 \caption{BLIP for (a) image captioning and (b) image-text matching.}
 \label{fig:blip}
\end{figure}

\textbf{Bootstrapping Language-Image Pre-training (BLIP)}~\cite{li2022blip}. 
Although there are many pre-trained language-image models, we focus specifically on BLIP~\cite{li2022blip} in this work, as it is one of the latest and represents the state-of-the-art (SOTA) till the beginning of this work. In Fig.~\ref{fig:blip}, we explain two functions that BLIP can perform, which are a subset of its capabilities. To get full details of the model, we refer readers to~\cite{li2022blip}.

\begin{itemize}
    \item The first one is caption generation, which is administered through a language modeling (LM) loss during training (Fig.~\ref{fig:blip}a). Given an image, the visual encoder of BLIP, a ViT, first decomposes it into $p^2$ patches and encodes each into a $d$-dimensional vector. The resulting $p^2{\times}d$ matrix is then passed to the text decoder (the LM part) for caption generation. Besides the patch embeddings, the text decoder also takes a prompt (e.g., ``\texttt{a picture of}'') as the beginning of the generated caption and it performs next token prediction recursively until reaching the termination criteria.

    \item The second function is image-text matching (ITM, Fig.~\ref{fig:blip}b), which outputs a probability value indicating if an image and a caption match with each other or not. Through the ViT, the image is encoded into $p^2$ $d$-dimensional embeddings. Meanwhile, the caption (with $t$ words) is also encoded into embeddings but through a bi-directional self-attention component. The cross-attention between the embeddings of the $p^2$ patches and $t$ tokens is then learned, resulting in a $p^2{\times}t$ matrix. Each element of this matrix reflects how strongly a word token attends to an image patch. This cross-attention matrix is our analysis focus while investigating the association between words and image patches.
\end{itemize}

BLIP has been pre-trained over hundreds of millions of image-text pairs and been successfully used in many downstream tasks. We use its pre-trained parameters for the above two functions in this work.
\setlength{\belowcaptionskip}{-10pt}
\setlength{\belowcaptionskip}{0pt}

\textbf{Segment Anything Model (SAM)}~\cite{kirillov2023segment}.
Our work also employs SAM to extract semantically meaningful segments from images. We use this model as a black box, where the input is an image and the outputs are multiple segments of the image, each is denoted as a binary mask indicating what pixels belong to the segment and what pixels do not. Despite SAM, we have also considered using region proposal or superpixel algorithms~\cite{ren2015faster, vedaldi2008quick, zhao2021human} to decompose an image. However, compared to SAM, those methods either more severely break the image semantics or contain more overlaps in the resulting segments. Thus, we chose SAM, which is also the SOTA method for image segmentation.

%% file: tex/4requirements.tex
\section{Goals and Tasks}
\label{sec:requirement}

\subsection{Goals}
By reviewing the literature and discussing with multiple domain practitioners working with pre-trained language-image models, we identified several pain points when model designers train, evaluate, and use these models. For example, Bertucci et al.~\cite{bertucci2022dendromap} elaborated the need of comprehending images before using them to train ML models. Gou et al.~\cite{gou2020vatld} detailed the need of assessing ML models to reveal their weaknesses. Multiple works~\cite{yang2020interactive, ming2019protosteer,chen2021towards, yang2022diagnosing} demonstrated the importance and benefit of allowing users to steer ML models. From the review and discussions, we scope our work to accomplish the following three goals.

\goalbox{G1} \textbf{Efficiently explore large-scale image datasets and expose potential data bias.} Prior to commencing the training of an image classifier (or other ML models), it is customary for domain practitioners to develop a comprehensive understanding of the image dataset and be mindful of potential issues associated with the dataset. For instance, as demonstrated in LIME~\cite{ribeiro2016should}, \texttt{snow} often presents in the training images of \texttt{wolf} as background. This undesired feature co-occurrence can mislead trained ML models into associating the existence of \texttt{snow} with \texttt{wolf}, a correlation that is, in fact, not always held. Consequently, the classifier might mislabel a \texttt{dog} image as \texttt{wolf} as long as \texttt{snow} is detected. However, due to the substantial volume of images involved, manually scrutinizing each image and subsequently extracting a comprehensive overview of the dataset poses a considerable challenge.

\goalbox{G2} \textbf{Reveal image features/objects that are not well covered by captions generated from ML models.} 
Despite the remarkable captioning performance exhibited by pre-trained language-image models, there remain cases where the models fail to generate captions for specific visual features within an image. For instance, in Fig.~\ref{fig:blip}, the sunglasses were not in the generated caption, despite being conspicuous features. By collecting these cases and summarizing the recurring features that frequently receive inadequate coverage, it becomes possible to unveil behavioral tendencies and potential limitations of the models. These limitations might encompass a deficient grasp of associations between certain visual traits and corresponding linguistic expressions. However, simultaneously examining a substantial volume of images along with their captions, while also identifying the prevalent inadequately covered features, proves to be an exceedingly complex endeavor.

\goalbox{G3} \textbf{Steer caption generation with human knowledge.} Another pain point faced by users of image captioning models is the limited control over the generation process. Often, the generated captions fail to adequately describe the visual features that are of particular interest to the users. Although those features may have been learned by the pre-trained models, certain constraints, such as caption length, can lead to the omission of the features' description. This occurs because the model deems them less crucial. However, the features that the model deems as important might not align with the features that hold significance in the users' perspective. To mitigate this, we target to enable users to actively guide the caption generation, i.e., directly specify the features of interest and steer the model to generate descriptions for them.

\subsection{Tasks}

Through iterative refinements over different design considerations, we have concretized our three goals into the following actionable tasks:

\taskbox{T1} \textbf{Textualize image contents and reveal the domain features and feature co-occurrence.} Thanks to the remarkable performance of image captioning models, the generated captions can capture the main contents of images very well. A holistic overview of enormous number of images can therefore be obtained from the captions without manually eyeballing individual images. The exploration of captions is also much easier than images, as they can be easily broken down. The appearing frequency of different words in the captions and their co-occurrence frequency effectively reflect the dominant and co-occurred image features. Accomplishing this task can substantially support \goalbox{G1}.

\taskbox{T2} \textbf{Segment image features and group them by their semantic similarity and received attentions.} An image may come with diverse features, e.g., a photo of a park may include several persons, their pets, the sun, and trees. An effective overview of an image dataset and its dominant features, therefore, requires to segment those features and cluster them to augment the data patterns (\goalbox{G1}).
The attention that different segments received when generating a word of the captions can also be used to curate the segments. This will help to disclose the association pattern between image features and caption words (\goalbox{G2}).

\taskbox{T3} \textbf{Coordinately explore the association between words (of captions) and visual features (of images).} A mapping between individual words and segmented image features is always necessary when exploring images through their captions or vice versa. This task enables the mutual exploration between the two data modalities (\goalbox{G1}). The correctness and strengths of the association will also reveal the quality of the pre-trained ML model (\goalbox{G2}), disclosing its potential weaknesses.

\taskbox{T4} \textbf{Empower users to steer caption generation by interacting with caption prompts or visual features.}
To achieve \goalbox{G3}, an interaction mechanism is needed to allow users to manipulate the settings before caption generation, engage with different image features, and compare the generated captions for feature verification.  This task aims to grant users greater control over the caption generation process, empowering them to craft captions that align with their preferences.

%% file: tex/5system.tex
\section{Methodology and Visual Analytics System}
\label{sec:system}

We explain our methodology for achieving the three goals in the following subsections, which cover the efficient exploration of extensive images, disclosure of caption coverage, and guiding caption generation.

\subsection{Efficient Image Data Exploration}
\label{sec:image_exploration}

Our first research goal is to provide an approach to efficiently exploring large-scale image datasets (\goalbox{G1}). We believe the images' captions generated using pre-trained language-image models can help here, as they can very well describe the main contents within images.

\begin{figure}[tbh]
\centering
\includegraphics[width=\columnwidth]{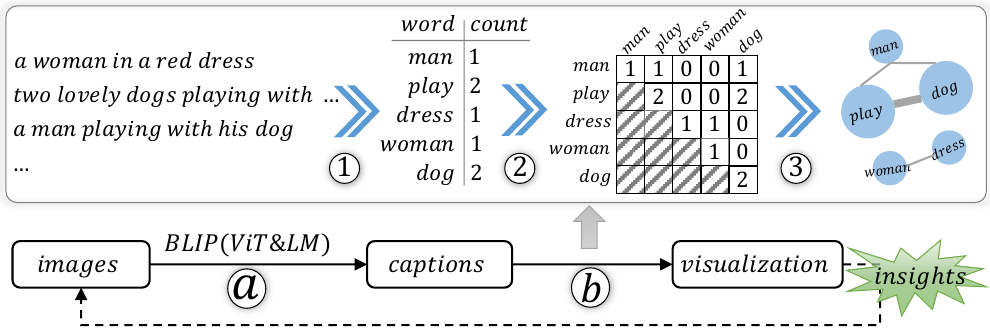}
\vspace{-0.2in}
\caption{Exploring images through our caption \graph{}.}
\label{fig:image_explore}
\end{figure}

Fig.~\ref{fig:image_explore} shows an overview of our approach.
Given a large-scale image dataset, we first feed individual images into a pre-trained language-image model for caption generation (Fig.~\ref{fig:image_explore}a). Specifically, the model here is the ViT and LM part of BLIP (Fig.~\ref{fig:blip}). The generated captions capture the essence of the images. An effective exploration of them can then provide insights into the original images without manually eyeballing individual ones. 
The transformation from images to texts is notably less labor-intensive, as the captioning process is mostly automated. Moreover, exploring those captions is more convenient compared to handling the images directly, as they can be easily deconstructed (into words with inherent semantics) and analyzed.

\setlength{\belowcaptionskip}{-5pt}
\setlength{\belowcaptionskip}{0pt}

We build a node-link diagram to effectively explore the generated captions (Fig.~\ref{fig:image_explore}b). The building process can be described as follows. \textbf{\textit{First}}, we tokenize the captions into words and standardize the same word in various formats (e.g., converting plural nouns into their singular form with TextBlob~\cite{textblob}). Using NLTK~\cite{bird2009natural}, we remove stop words, such as `a', `the', and `with', as they carry no semantics. For the remaining unique words, we count their appearance in the captions.
\textbf{\textit{Second}}, a co-occurrence matrix is constructed, each cell denotes the count that the words from the corresponding row and column occurred simultaneously in one caption. The resulting matrix will be symmetric and we only need half of it. \textbf{\textit{Third}}, a node-link diagram is built from the co-occurrence matrix. Each node represents a unique word and its size reflects the count that the word occurred in all captions (i.e., its frequency). A link between two nodes indicates the corresponding words co-occurred in at least one caption. Its width reflects the co-occurrence frequency. We use square root mapping and size clipping to handle too small/large nodes, as well as too thin/thick links.

The node-link diagram is visualized using the force-directed layout in the \graph{} view of our system (Fig.~\ref{fig:teaser}a). Dragging-and-dropping a node will pin it at a fixed position. Clicking a node will show details of the corresponding word and highlight image segments that have strong cross-attentions with it in the \scatterplot{} view (explained later). Nodes and links with very small values can be filtered out to help users focus on the more important nodes and links (i.e., the dominant visual features and their co-occurrence relationships). Interactively exploring the node-link diagram helps to comprehend the underlying images and image features. For example, Fig.~\ref{fig:teaser}a shows the node-link diagram built on top of the captions generated from the \texttt{tench} class images of ImageNet~\cite{imagenet}. From the visualization, it is evident that most of the images contain \texttt{fish} and \texttt{man}, and the two features appear concurrently in most images (reflected by the thick link between them). Considering the big node for \texttt{holding}, we can imagine that most images include the content of ``a man holds a fish''.

\begin{figure}
\centering
\includegraphics[width=0.8\columnwidth]{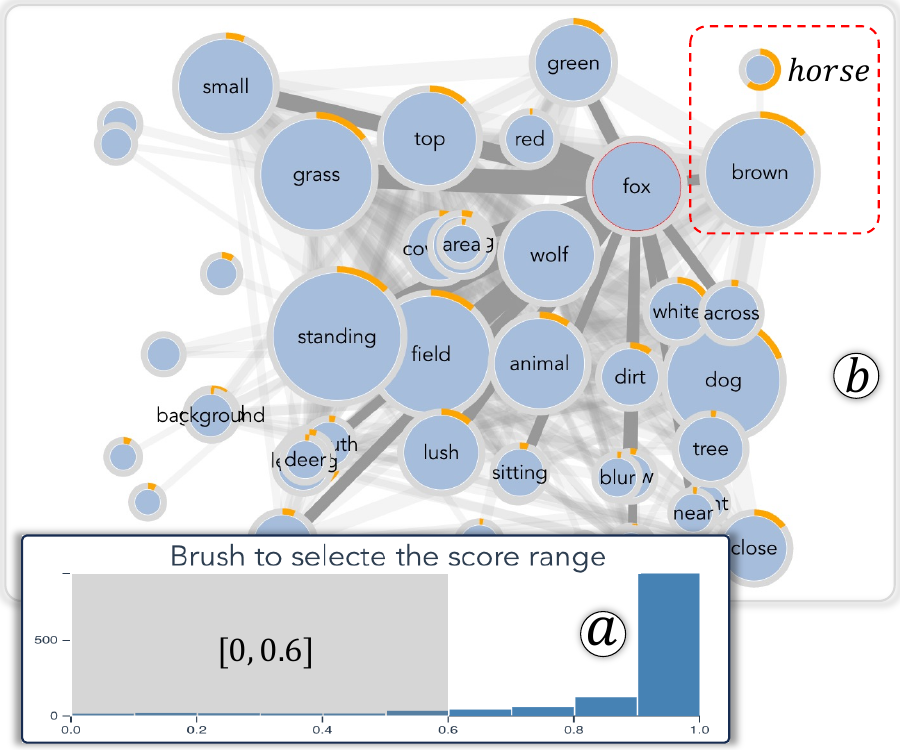}
\vspace{-0.1in}
\caption{The outer ring/arc of each node reflects the portion of the corresponding word in a selected ITM score range ([0, 0.6] in this case).}
\label{fig:score_distribution}
\end{figure}

Additionally, each image-caption pair is also fed into the ITM part of BLIP (Fig.~\ref{fig:blip}b) to evaluate the caption quality. The histogram in Fig.~\ref{fig:score_distribution}a (popped up after clicking the ``Score Distribution'' in Fig.~\ref{fig:teaser}-a2) shows the ITM score distribution of all image-caption pairs. 
To present the word distribution in a certain score range, we use the outer ring of a node to reflect the portion of the corresponding word in that range.
In Fig.~\ref{fig:score_distribution}a, the brushed range is $[0, 0.6]$, and the outer rings of all nodes in Fig.~\ref{fig:score_distribution}b show the portion of the corresponding words appearing in these low-quality captions.
Specifically, the explored images here are \texttt{red\_wolf} images from ImageNet and a big portion of the word \texttt{horse} appears in low-quality captions. The word also has a connection to \texttt{brown}, indicating a confusion between \texttt{red\_wolf} and \texttt{brown\_horse}. On the other hand, as the \texttt{horse} node is small, the issue is less severe.

Note that we have also considered other design choices, such as using a bar chart to rank the words, an arc diagram to connect them, or a chord diagram to reflect their frequency. These designs, however, are less space-efficient than a node-link diagram and our current visualization is also more intuitive for domain practitioners to interact with. 

\setlength{\belowcaptionskip}{-10pt}
\setlength{\belowcaptionskip}{0pt}

\subsection{Image-Text Association and Caption Coverage}

To accomplish \goalbox{G2} of revealing caption coverage over image features, we need to first extract meaningful image features (\taskbox{T2}), and then, disclose their association with different words inside the captions (\taskbox{T3}).

For Task \taskbox{T2}, we employ the pre-trained SAM~\cite{kirillov2023segment} to automatically extract semantic segments out of each image. The resulting segments could be very fragmentary or have big feature overlaps, and thus we perform some filtering steps to distill the most representative ones. Specifically, we first exclude segments that are too small to carry any semantics. These are the segments whose size (number of pixels) is less than 1\% of the corresponding image. This step significantly reduces the number of segments, i.e., $66\%{\sim}88\%$ of the total for different classes. Second, we compute the intersection-over-union (IoU) over all pairs of segments from the same image. For any pair, if the IoU is over 0.85 (i.e., having a big overlap), we keep only one out of the two to reduce redundancy. This further reduces the number of segments by $1\%{\sim}4\%$ of the total across different classes. The choices of the two values, i.e., 1\% and 0.85, are based on our empirical studies, but can be altered on-demand during exploration of different datasets.

To cluster the segments for efficient visual summary and exploration, we feed all filtered segments into the ViT of BLIP to generate their high-dimensional embeddings. These embeddings are then projected to 2D through a tSNE+scatterplot visualization (i.e., the \scatterplot{} view, Fig.~\ref{fig:teaser}b). Segments representing similar features from different images are close to each other in the 2D space, facilitating the identification of common visual features. Fig.~\ref{fig:case_tench_weakness}a (as well as Fig.~\ref{fig:teaser}b and Fig.~\ref{fig:case_tench}) shows this scatterplot visualization. By lasso-selecting different clusters, we can easily explore similar segments, such as the \texttt{fish} and \texttt{face} segments in Fig.~\ref{fig:case_tench_weakness}b and~\ref{fig:case_tench_weakness}c, respectively. Clicking a point will pop-up a window showing the corresponding segment, the original image that the segment resides, and the generated caption (Fig.~\ref{fig:teaser}-b3).

\setlength{\belowcaptionskip}{-10pt}
\begin{figure}[tbh]
\centering
\includegraphics[width=\columnwidth]{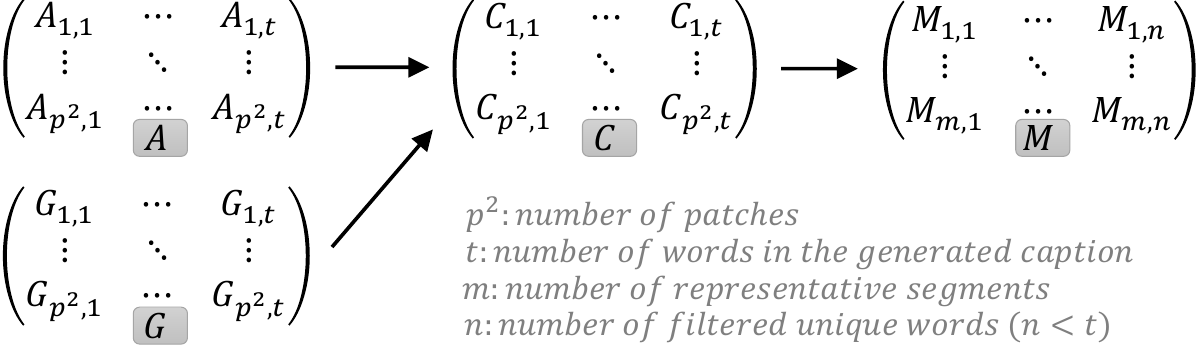}
\vspace{-0.2in}
\caption{The Grad-CAM computation for a single image-caption pair.}
\label{fig:cross}
\end{figure}
\setlength{\belowcaptionskip}{0pt}

For Task \taskbox{T3}, we associate the words from captions with the segments from images through Grad-CAM~\cite{selvaraju2017grad}, as it has been widely adopted and proven to be effective in early works~\cite{li2021align,li2023lavis}. Fig.~\ref{fig:cross} illustrates the computation process. 
\textbf{\textit{First}}, for a single image-caption pair, we get its $p^2{\times}t$ cross-attention, denoted as $A$, from the ITM part of BLIP (Fig.~\ref{fig:blip}b). Each cell of this matrix reflects the attention between the corresponding pair of image patch (row) and caption word (column). \textbf{\textit{Second}}, we get the gradient of $A$ with respect to the ultimate ITM score, and name it as $G$ (i.e., $G_{i,j}{=}\frac{\partial ITM}{\partial A_{i,j}}$). These two matrices are then merged through an element-wise multiplication to get $C$, which is the Grad-CAM. $C$ is better than $A$ in reflecting the association between words and patches, as there are cases where the attention is strong between a patch and a word but the gradient is close to 0. \textbf{\textit{Third}}, we remove the columns of $C$ for stop words and aggregate the rows of $C$ based on how different patches form into segments. The resulting matrix $M$ has a size of $m{\times}n$, where $m$ is the number of representative segments and $n$ is the number of unique words for the given image-caption pair. Each element of this matrix denotes the attention between a word and a segment. \textbf{\textit{Lastly}}, the computation is extended to all image-caption pairs, and the resulting $M$ from them are joined together by computing the union of their rows and columns to generate $M^{\cup}$.

\textit{
We want to further clarify two details when generating $C$ in Fig.~\ref{fig:cross}. First, there are multiple copies of $A$, one for each attention head of the BLIP. In our case, the underlying Transformer of BLIP has 12 layers, each layer has 12 heads. So, we have 144 copies of $A$ in total. Similar to previous works~\cite{li2021align}, we aggregate the 12 $A$ from the same layer and use RefCOCO+~\cite{kazemzadeh2014referitgame, li2021align} \footnote{RefCOCO+ can be considered as a ground-truth dataset, in which what patches each word should attend to have been well annotated. It can be used to evaluate the association quality learned by each layer/head of the BLIP model. We have included details of the grounding experiments in our Appendix.} to perform a grounding study to get the layer that has the best image-text grounding quality. In our case, the best layer is the 7th layer of BLIP. We use this layer's cross-attention by default, but also allow users to switch to other layers for exploration (through the widget in Fig.~\ref{fig:teaser}-b2). Second, there are also some image resizing operations when computing the Grad-CAM of images, as the images may not be in a square shape. Grad-CAM is resized into the original image size and pixel-wise accumulation is conducted to generate the Grad-CAM for each patch (see our Appendix for these details).
}

\textbf{\textit{Each column of $M^{\cup}$}} reflects the attention strengths between a single word and all segments. We use these values to color the points in the \scatterplot{} view when a word is selected. For example, in Fig.~\ref{fig:teaser}a, the word \texttt{fish} is selected, its attentions to all segments are shown in the scatterplot of Fig.~\ref{fig:teaser}b, where darker orange indicates stronger attention. Clicking a point shows the segment, and the spatial distribution of the selected word's Grad-CAM over the segment will be superimposed as a heatmap (red$\rightarrow$blue: strong$\rightarrow$weak). For example, in Fig.~\ref{fig:teaser}-b3, the head region of the extracted fish segment received most of the attention from the word \texttt{fish}. The segments selected through lasso selections also have this overlapped heatmap, as shown in Fig.~\ref{fig:teaser}-b1 and Fig.~\ref{fig:case_tench}.

If a segment is among the top-$k$ ($k{=}$3 by default) segments that received the most attention from a word, we say the segment is \textit{covered} by the word. Iterating through all words, we get a coverage number for the segment, reflecting how many words strongly attend to it. The scatterplot in the \scatterplot{} view can also be colored by individual segments' coverage. As shown in Fig.~\ref{fig:case_tench_weakness}a, segments with increasing coverage are mapped to points from yellow to green. 
The green segments are covered by multiple words and are often the dominant features. For example, the green cluster of fish segments in Fig.~\ref{fig:case_tench_weakness}b are covered by the words \texttt{fish}, \texttt{large}, and \texttt{swimming}.
In contrast, the yellow segments are not sufficiently covered by any caption words, which deserve more attention for caption quality improvement.

\textbf{\textit{Each row of $M^{\cup}$}} reflects the attention that the corresponding segment has received from all caption words. When a segment is selected, the caption for its residing image will be shown and we use yellow background to highlight the words where the segment received most of its attentions from. For example, in Fig.~\ref{fig:teaser}-b3, the selected segment received attentions mainly from the words \texttt{large} and \texttt{fish}. The word \texttt{fish} is in red, indicating it is the currently selected word from Fig.~\ref{fig:teaser}a.

\subsection{Steering Caption Generation}
The LM part of BLIP generates captions based on two inputs (Fig.~\ref{fig:blip}a), the starting prompt and image patches' embeddings. Therefore, without updating the pre-trained BLIP, we can accomplish \goalbox{G3} of caption steering by injecting human knowledge into these two inputs. 

\setlength{\belowcaptionskip}{-10pt}
\begin{figure}[tbh]
\centering
\includegraphics[width=\columnwidth]{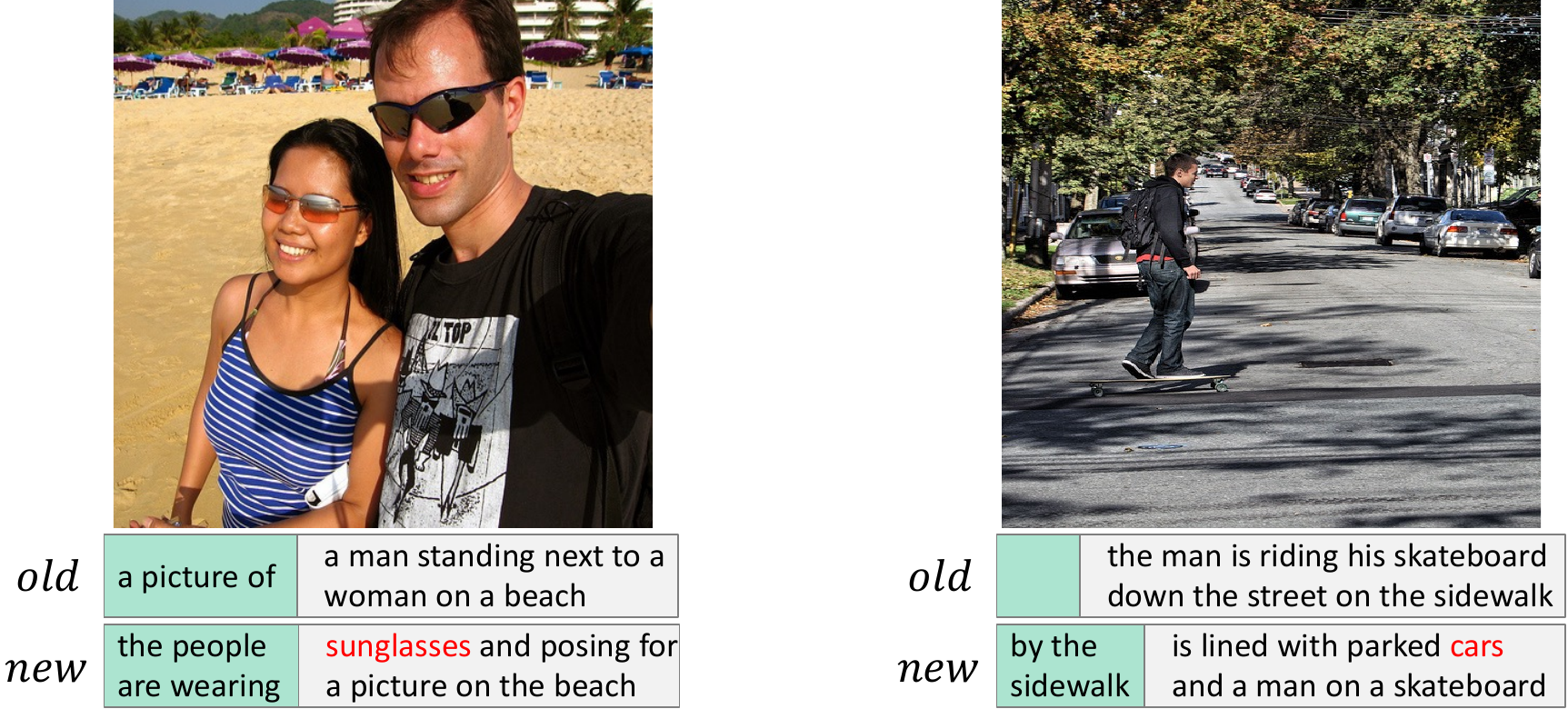}
\vspace{-0.2in}
\caption{Steering caption generation with guiding prompts.}
\label{fig:prompt_example}
\end{figure}
\setlength{\belowcaptionskip}{0pt}

\textbf{\textit{First}}, the starting prompt can guide the generation of certain words in the caption. General prompts, such as ``\texttt{a picture of}'' or an empty string, are often used by default, allowing the model to look at different features equivalently.
In cases where we want to generate captions for a specific feature, the prompt can be customized towards that feature to exploit the potential of the model. 
For example, in Fig.~\ref{fig:prompt_example}, the old captions generated with the general prompts discard the \texttt{sunglasses} (left) and \texttt{cars} (right) to compromise for other more conspicuous features. Inserting the verb ``\texttt{wearing}'' and location descriptor ``\texttt{by the sidewalk}'' into the respective prompts (texts in green) leads the model to include these features back in the captions. This is also a proof that the pre-trained BLIP was able to recognize these features beforehand, but considered them less important in the respective images.

\setlength{\belowcaptionskip}{-10pt}
\begin{figure}[tbh]
\centering
\includegraphics[width=\columnwidth]{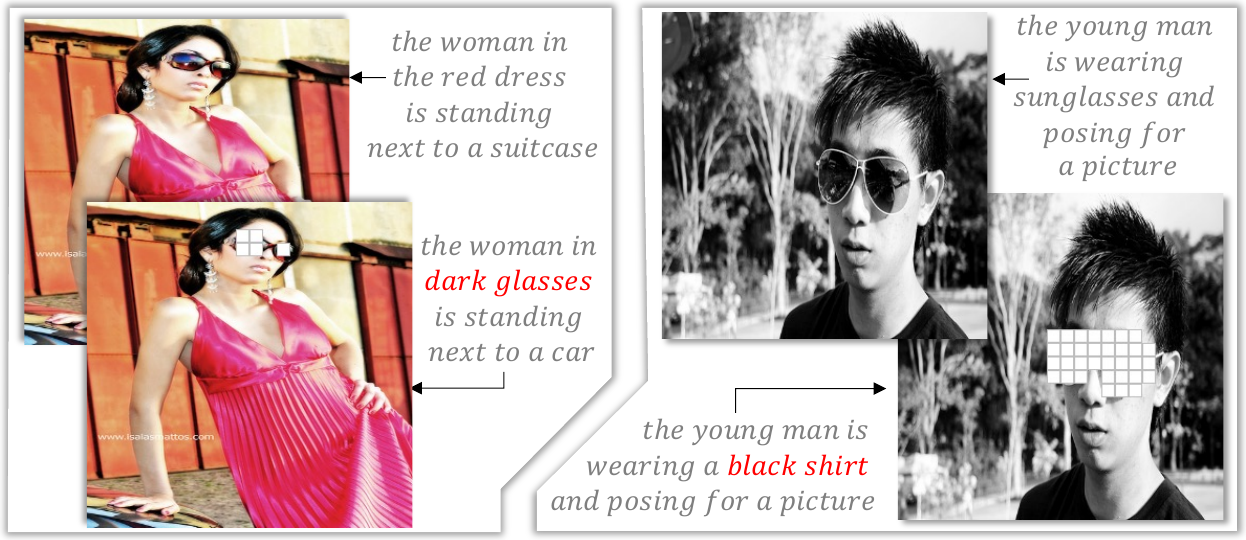}
\vspace{-0.2in}
\caption{Steering caption generation by directly emphasizing (left, weight =5) or de-emphasizing (right, weight=0) the \texttt{glasses} patches.}
\label{fig:steer_glass}
\end{figure}
\setlength{\belowcaptionskip}{0pt}

\textbf{\textit{Second}}, we can also steer the caption generation process by directly emphasizing or de-emphasizing different patches. As shown in Fig.~\ref{fig:blip}, the input image is decomposed into $p^2$ patches, each is encoded into a $d$-dimensional vector through the ViT. To guide the BLIP (LM) model to generate captions for certain visual features, we can force the model to look at them by increasing the corresponding patches' weight. On the contrary, weights can also be decreased for patches with negligible features. Specifically, we create a $p^2$ matrix, all elements of which are initialized to 1, and overlay this matrix on top of the image of interest. 
In Fig.~\ref{fig:steer_glass}, two examples show the results of emphasizing and de-emphasizing the \texttt{sunglasses} feature, respectively. On the left, the caption for the original image does not contain any descriptions of the \texttt{sunglasses}. After scaling up the weights of sunglasses' patches (white squares with gray borders) from 1 to 5, we get a new caption that includes the desired words \texttt{dark glasses}. 
To reverse this process, on the right, the original caption of the selected image involves \texttt{sunglasses}. After scaling down the weights to 0 for the corresponding image patches, the word \texttt{sunglasses} disappears. Instead, the words \texttt{black shirt} appear, as the patches for the shirt region receive relatively more attentions shifted from the \texttt{sunglasses} pixels.

The \steer{} view in Fig.~\ref{fig:teaser}c provides an interface to directly apply the above two manipulations. Users can directly type a prompt in Fig.~\ref{fig:teaser}-c2 as the starting prompt. By clicking a pixel of the selected image in Fig.~\ref{fig:teaser}-c1, the patch that the pixel resides in will be selected and marked as white. Users can then increase or decrease the weight of the corresponding patches. After clicking the ``Generate'' button, a new caption will be generated with the specified prompt (texts in green) or weights applied to the selected patches (Fig.~\ref{fig:teaser}-c3).

Note that our caption steering solutions have certain limitations that should be acknowledged.
First, we assume the pre-trained captioning model is able to recognize the desired features. Based on this, we guide the model to focus on the desired features and generate descriptions for them. However, if this assumption does not hold, the model can never generate desired captions without further fine-tuning.
Second, the patches' weight in the second approach is an important hyperparameter, but we currently can only get its value through a trial-and-error process and an overlarge weight will disrupt the caption's semantics. We have investigated several ways to fix this problem, such as normalizing the weight values or smoothing the $p^2$ weight matrix (with a Gaussian smooth). These optimizations brought some improvements in different cases but no consistent solutions have been established. Our domain experts also found this part particularly intriguing and have initiated some theoretical analyses towards a more robust solution.

%% file: tex/6casestudy.tex
\section{Case Study, Findings, and Experts' Feedback}
\label{sec:case_study}
In this section, we demonstrate how our system can be used to explore large-scale image datasets, detect insufficiently covered image features from captions, and steer caption generations. 
These cases are explored together with five deep learning experts, all are full-time research scientists with special interests in Transformers, model pre-training, and multi-modal learning. The experts all have $5{\sim}10$ years of experience in deep learning and have worked on various vision and language models, with multiple publications in the related fields. 
We studied our system with them following the protocol of \textit{guided explorations + think-aloud discussions} in multiple exploration sessions.
Each of these sessions extended over a span of $1{\sim}2$ hours, during which, we provided comprehensive explanations on different system components and walked the experts through various cases. The experts were also encouraged to engage with our system, share their insights on our approach, and offer recommendations for enhancements.
At the end of the explorations, we conducted open-ended interviews with them to collect their feedback and suggestions (Sec.~\ref{sec:feedback}). Two image datasets have been used over the studies: ImageNet~\cite{imagenet} and COCO~\cite{coco}.

\subsection{Large-Scale Image Dataset Exploration}
ImageNet~\cite{imagenet} is often used to train image classifiers. To better differentiate different classes, it is crucial to know the dominant features within individual classes and potential confusion between classes. Therefore, we explore this image dataset class-by-class. 

\begin{figure*}[tbh]
\centering
\includegraphics[width=\textwidth]{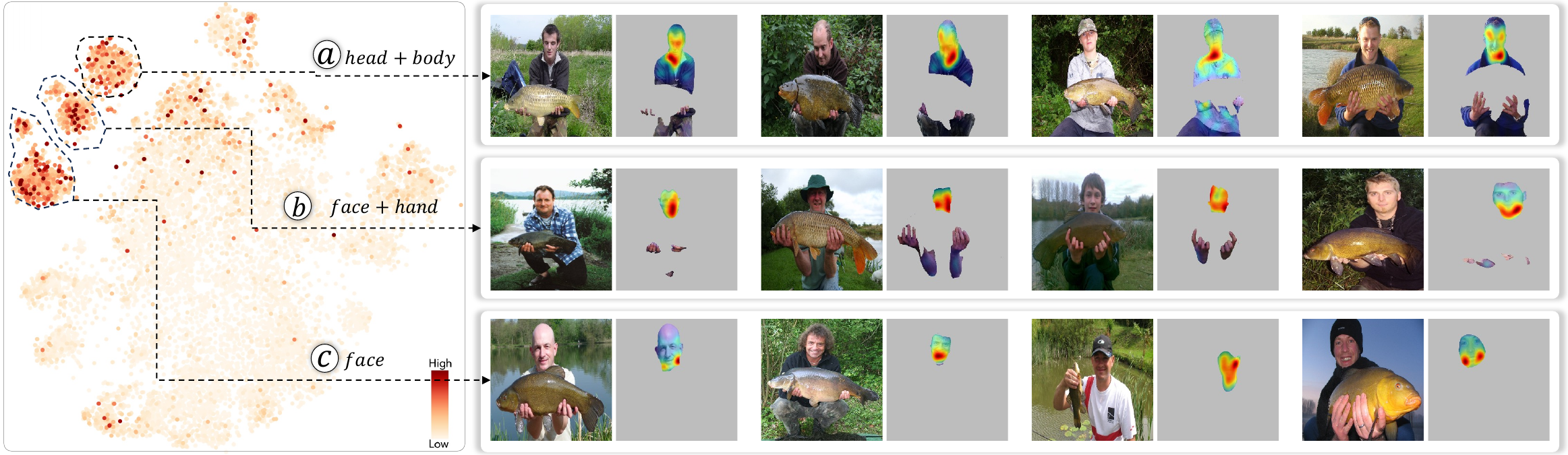}
\vspace{-0.25in}
\caption{For \texttt{tench} images, \texttt{man} is a dominant feature. The three dark orange clusters show three groups of segments that the word \texttt{man} strongly attends to. Some randomly selected segments are visualized on right. The overlaid heatmaps help to disclose the spatial distribution of the attentions.}
\label{fig:case_tench}
\end{figure*}

From the header of the \graph{} view, we picked the \texttt{tench} class, as it has been well studied in early works~\cite{hohman2019s} and some known data issues have been reported, i.e., images of this class often come with a person holding a tench fish. Starting with this class also works as a sanity check to assess our system's ability to identify previously reported issues.
From the visualization in Fig.~\ref{fig:teaser}a, we have several observations. 
\textbf{\textit{First}}, the word occurred the most in the generated captions is \texttt{fish}, instead of \texttt{tench}. This is expected by the experts as the pre-trained BLIP is not able to differentiate different types of fish. 
\textbf{\textit{Second}}, from the nodes' sizes and links' widths, we can easily identify the dominant and co-occurred features in this class, i.e., \texttt{man}, \texttt{hand}, \texttt{fish}, and \texttt{holding}. This effectively highlights the potential issue of the images (i.e., undesired feature co-occurrence). People should be aware that ML models trained on these images will heavily rely on the presence of \texttt{man} and \texttt{hand} to recognize an image as a \texttt{tench} image. 
\textbf{\textit{Third}}, Fig.~\ref{fig:teaser}-a1 also reflects that \texttt{man} occurs much more frequently than \texttt{woman} and \texttt{person}, suggesting a gender skew in fishing activities. Consequently, models trained on these images may involve potential gender bias as well. 
\textbf{\textit{Lastly}}, exploring different words in the \graph{}, we can see the corresponding visual features are very well clustered in the \scatterplot{} view. For example, the dark orange cluster in Fig.~\ref{fig:teaser}b corresponds to segments strongly associated with the currently selected word, i.e., \texttt{fish}. Similarly, in Fig.~\ref{fig:case_tench}, the three dark orange clusters show the segments that the word \texttt{man} strongly attends to, i.e., \texttt{head+body}, \texttt{face+hand}, and \texttt{face}. 
The heatmaps over individual segments further enrich our understanding. For example, although the segments in Fig.~\ref{fig:case_tench}b are for \texttt{face+hand} pixels, the word \texttt{man} mostly attends to \texttt{face} pixels.
These coordinated explorations validate the association between different caption words and image features, enhancing users' comprehension of the images.

\setlength{\belowcaptionskip}{-10pt}
\setlength{\belowcaptionskip}{0pt}

\setlength{\belowcaptionskip}{-10pt}
\begin{figure}[tb]
\centering
\includegraphics[width=\columnwidth]{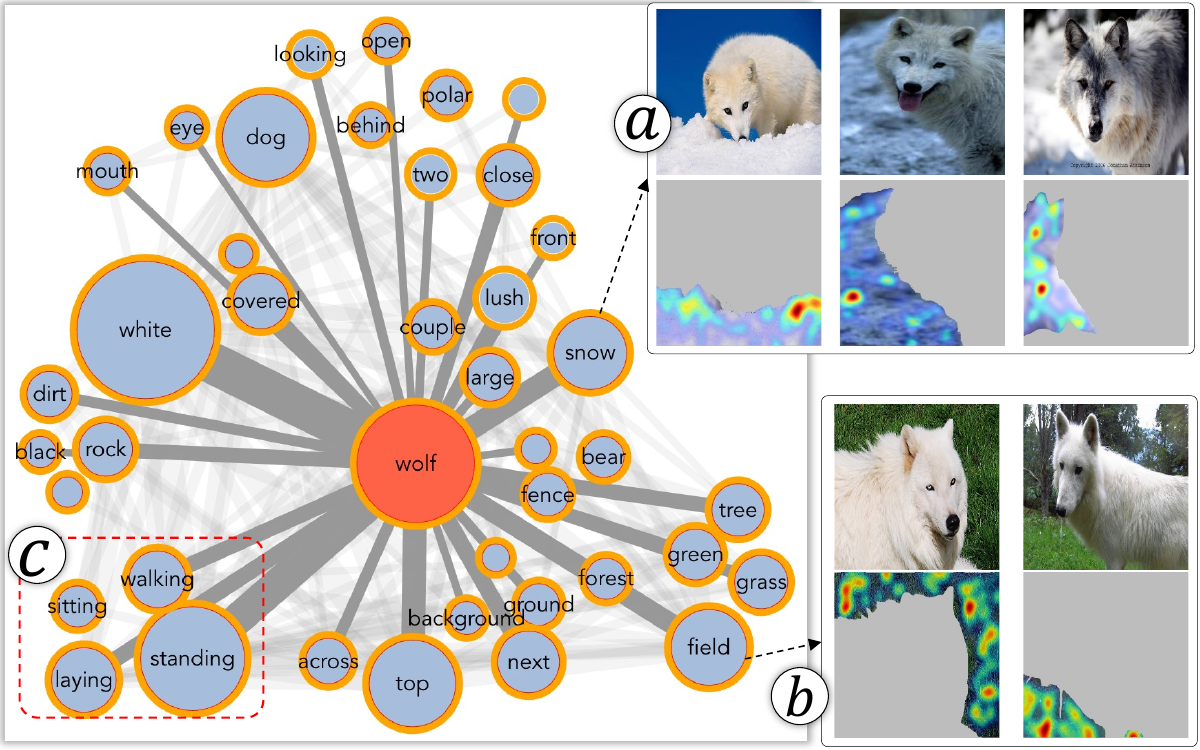}
\vspace{-0.25in}
\caption{\texttt{Snow} and \texttt{field}  frequently co-occurred with \texttt{white\_wolf}.}
\label{fig:case_wolf}
\end{figure}
\setlength{\belowcaptionskip}{0pt}

Next, we switched to the \texttt{white\_wolf} class of ImageNet. In the LIME paper~\cite{ribeiro2016should}, the authors intentionally curated a small number of \texttt{wolf} images with snowy backgrounds to explain how undesired feature co-occurrence could lead to a biased model. Inspired by their work, we want to see how severe the co-occurrence between \texttt{white\_wolf} and \texttt{snow} is in this real-world dataset.
As shown by the node-link diagram in Fig.~\ref{fig:case_wolf}, \texttt{snow} indeed frequently occurred in \texttt{white\_wolf} images, reflected by the corresponding node's size. Clicking on the node of \texttt{snow}, we can see a clear cluster of snow segments in the \scatterplot{} view (not shown here) responding strongly to the word. Fig.~\ref{fig:case_wolf}a shows some example segments and the corresponding images.
Apart from \texttt{snow}, our system also detects other co-occurred features that are frequently present in the natural habitat of \texttt{white\_wolf}. For example, the node for \texttt{field} is also large. As shown in Fig.~\ref{fig:case_wolf}b, this node, along with several other nearby nodes (e.g., \texttt{green} and \texttt{grass}) discloses other high-frequent background features in \texttt{white\_wolf} images.
Furthermore, the dominant words from the captions also capture the main poses of wolves in the images (Fig.~\ref{fig:case_wolf}c), such as \texttt{standing}, \texttt{laying}, and \texttt{walking}. The overview from our system provides a holistic picture for the dominant postures of wolves and their living environment. 

\begin{figure}[tb]
\centering
\includegraphics[width=\columnwidth]{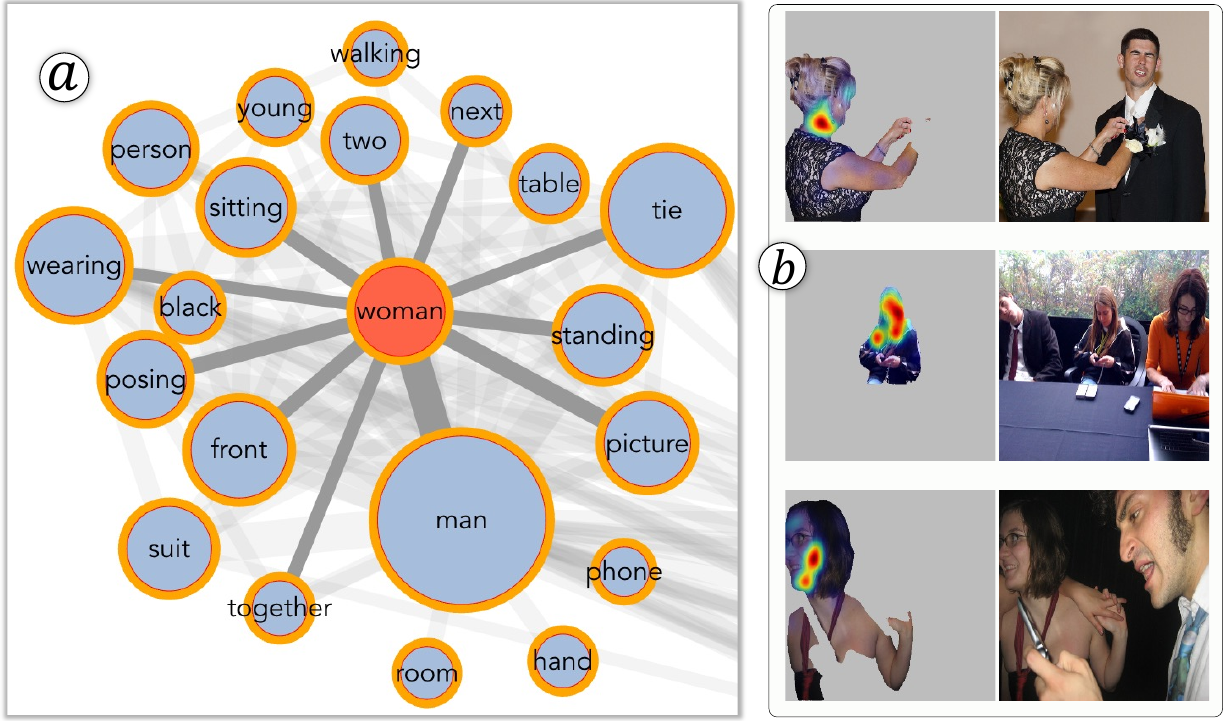}
\vspace{-0.25in}
\caption{Exploring COCO images with both \texttt{person} and \texttt{tie} objects.}
\label{fig:case_tie}
\end{figure}

The \graph{} also helps to structurally explore different image features and disclose their imbalanced distributions. We use the COCO dataset~\cite{coco} to demonstrate this. COCO contains images with well-annotated objects in 80 categories, and each image often involves multiple objects. While exploring the images with both \texttt{person} and \texttt{tie} objects, we found the word \texttt{man} is a high-frequent feature (i.e., the biggest node in Fig.~\ref{fig:case_tie}a) whereas \texttt{woman} has a much lower frequency. Examining the co-occurrence features of \texttt{woman}, we found it has a thick link to \texttt{man} but a very thin link to \texttt{tie}. This reveals two insights: (1) men occur much more frequently than women in the \texttt{person-tie} images, and (2) the occurrence of women is more often accompanied by men than by wearing ties. By examining segments that strongly respond to \texttt{woman} and checking the corresponding images in the \scatterplot{} view, we verified these insights (Fig.~\ref{fig:case_tie}b). By structurally exploring the objects within images, we reveal the imbalanced distribution of genders and their co-occurrence with \texttt{tie}. Such an imbalanced distribution might lead to a trained ML model failing to detect ties worn by women, and thus deserves more attention.

More exploration results and discovered insights have been included in our Appendix. The knowledge on the frequency of different image features and their co-occurrence provides valuable insights into a dataset, helping to comprehend the dataset before using it for any model training. Although whether a feature co-occurrence should be considered as a data bias or not is very application-specific, it is crucial to reveal it to model designers to support their decision-making. 
The awareness of this information also helps them select datasets and remain vigilant regarding model issues stemming from the data. This effort closely aligns with the growing emphasis on data-centric AI~\cite{strickland2022andrew,wang2023visual}.

\subsection{Caption Coverage and Model Behavior Pattern}
\label{sec:weakness}

While the image captioning model exhibits remarkable performance, it is not without imperfections.
Revealing what common image features that the model often overlooks can shed light on its weaknesses and offer valuable insights for its enhancement. Our system can effectively disclose this, based on the feature coverage of the generated captions. To demonstrate this capability, we continue our class-level explorations, as images of the same class often share many features.

\begin{figure*}[tbh]
\centering
\includegraphics[width=\textwidth]{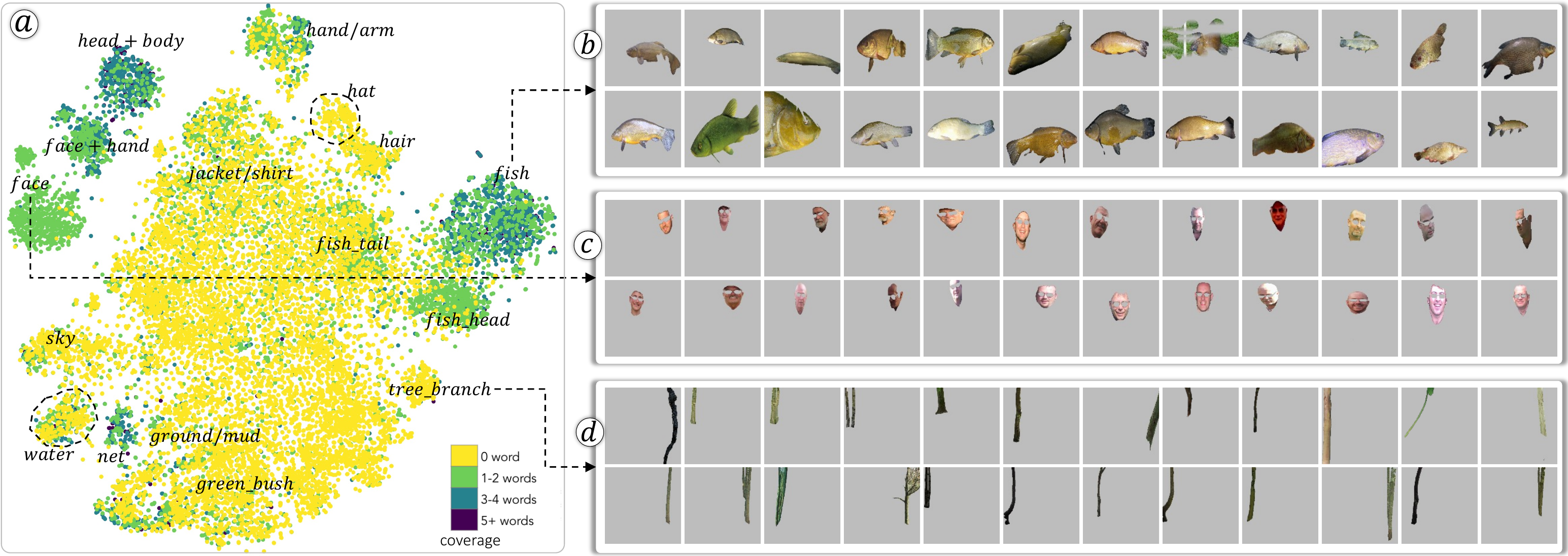}
\vspace{-0.2in}
\caption{(a) The word coverage over segments extracted from \texttt{tench} images. (b, c, d) Three example clusters are explored through lasso selections.}
\label{fig:case_tench_weakness}
\end{figure*}

Continuing the \texttt{tench} case, we feed all SAM extracted segments from the \texttt{tench} images into ViT and use tSNE to lay out their resulting embeddings in the \scatterplot{} view (Fig.~\ref{fig:case_tench_weakness}a). 
From the layout, we observed that the segments are very well clustered. By individually exploring the clusters through lasso selections, we could easily figure out the semantics of each cluster. For example, Fig.~\ref{fig:case_tench_weakness}b,~\ref{fig:case_tench_weakness}c, and~\ref{fig:case_tench_weakness}d show the segments of \texttt{fish}, \texttt{human\_face}, and \texttt{tree\_branch}, respectively.

\setlength{\belowcaptionskip}{-10pt}
\setlength{\belowcaptionskip}{0pt}

\setlength{\belowcaptionskip}{-10pt}
\setlength{\belowcaptionskip}{0pt}

The segments in Fig.~\ref{fig:case_tench_weakness}a are colored using their word coverage, i.e., the number of unique words strongly attending to the corresponding segments. The color from yellow to green reflects the increasing coverage (see the legend). 
Evidently, the segments related to \texttt{fish} (the two clusters on the right) and \texttt{man} (the three clusters on the top-left) received the most coverage. Apart from them, many points/segments are in yellow (i.e., zero word coverage).
If the yellow points are dispersed, such as the bottom right region of Fig.~\ref{fig:case_tench_weakness}a, the corresponding segments have fewer features in common (i.e., fragmented pieces). However, if the yellow points form into isolated clusters, it usually indicates that some common features are inadequately covered by the captions, such as the clusters for \texttt{hat}, \texttt{tree\_branch}, \texttt{water} and \texttt{sky} in Fig.~\ref{fig:case_tench_weakness}a. 

To verify this, we picked the \texttt{hat} cluster for detailed investigations. 
Fig.~\ref{fig:caption}a (the middle column) shows some randomly picked \texttt{hat} segments. Their residing images are shown in the left column. The generated captions (below the images) do not contain descriptions for the \texttt{hat}.
Consistent phenomena were observed from most images with this feature. This recurring behavior pattern, i.e., the model tends to overlook the \texttt{hat} feature, reveals a weakness of the model. Similarly, the segments of background \texttt{water} (the bottom-left cluster in Fig.~\ref{fig:case_tench_weakness}a) are also insufficiently covered, and some randomly picked image examples are shown in Fig.~\ref{fig:caption}b.
Although these features may appear less prominent than \texttt{fish} and \texttt{man}, they exist in a substantial number of images and may be of interest to users in certain scenarios. Nevertheless, given the limited control over the caption generation process, the pre-trained BLIP model struggles to generate descriptions for these features once it has decided to overlook them.

Similar caption coverage insufficiency has also been found from the COCO dataset. In this case, we caption images that contain both \texttt{car} and \texttt{person} objects. Although the \texttt{car} object is known to exist in all images, the generated captions cannot cover it very well, as reflected by the yellow cluster of \texttt{car} segments from the \scatterplot{} view (not shown here for space limit). 
As some randomly picked examples shown in Fig.~\ref{fig:caption}c, this omission is due to the \texttt{car} being in the background or outsized by other objects. Although the \texttt{cars} do not play key roles in these images, they (along with \texttt{person}) are the focus of captioning in this case, and users would expect to see them in the generated captions.

\subsection{Steering Caption Generation}

After the overlooked visual features have been identified, users can decide if they want to update the captions to include those features or not (based on their needs). If yes, they can steer the caption generation process using the \steer{} view of our system in two ways.

First, users can update the starting prompt to lead the model toward the desired features. As shown in Fig.~\ref{fig:teaser}c, a \texttt{tench} image containing the \texttt{hat} segment in the yellow cluster of Fig.~\ref{fig:case_tench_weakness}a is selected. The original caption, generated with the prompt ``\texttt{a picture of}'', does not contain descriptions for the \texttt{hat} feature. However, after modifying the prompt to ``\texttt{the person is wearing}'', we successfully guided the model to generate a caption with the word \texttt{hat}. The new prompt can also be applied to all images containing \texttt{hat} segments for a batch processing. For example, in a lasso selection of 214 images containing segments in the \texttt{hat} cluster of Fig.~\ref{fig:case_tench_weakness}a, 193 newly generated captions contain the word \texttt{hat}/\texttt{beanie}/\texttt{hoodie}, resulting in a success rate of 90.19\%. For the 21 failed cases, the words \texttt{jacket}, \texttt{gloves}, and \texttt{goggles} primarily respond to the word \texttt{wearing} in our prompt. For these cases, we have to explicitly emphasize the \texttt{hat} pixels to better steer the model. Also, 9 out of the 21 images indeed have no \texttt{hat} features, but the SAM extracted segment is very similar to a \texttt{hat}. We have included the full details of this batch processing experiment in our Appendix.

\setlength{\belowcaptionskip}{-10pt}
\begin{figure}[tbh]
\centering
\includegraphics[width=\columnwidth]{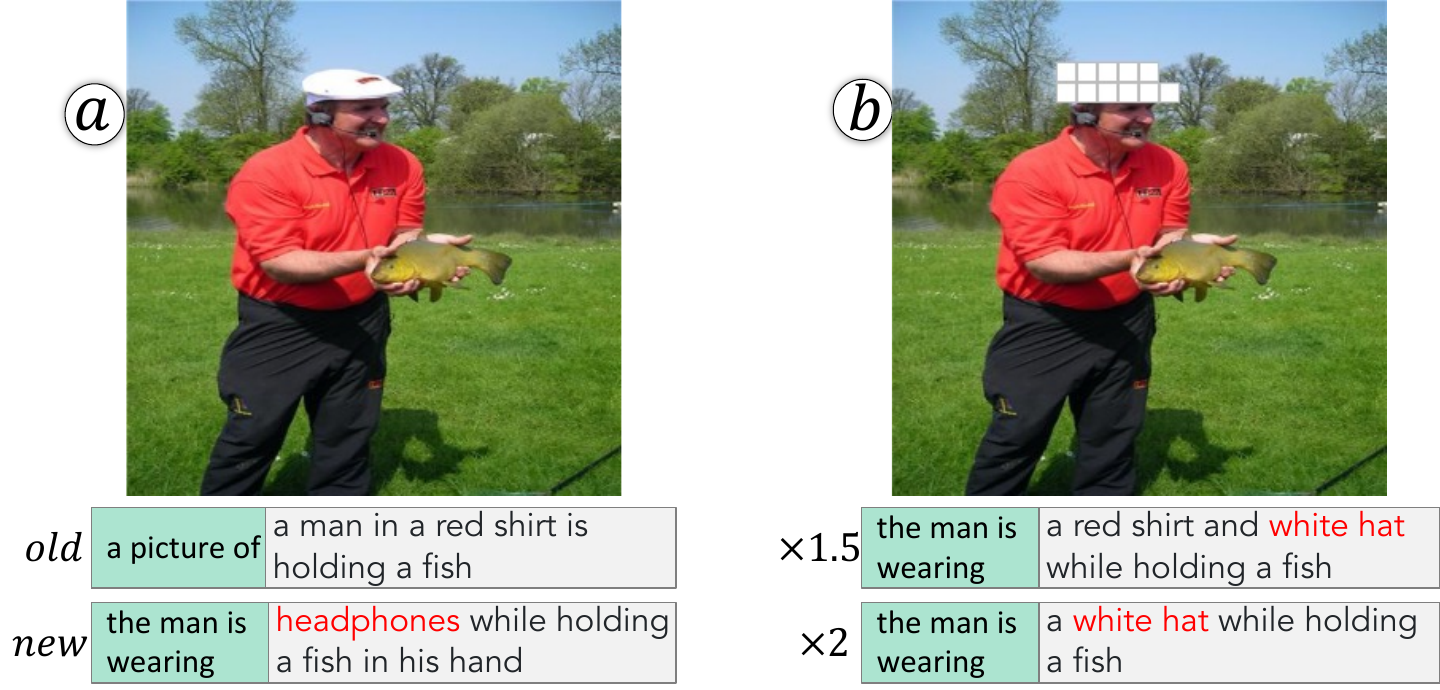}
\vspace{-0.2in}
\caption{Iteratively steering the model to generate descriptions for \texttt{hat}.}
\label{fig:caption_case}
\end{figure}
\setlength{\belowcaptionskip}{0pt}

\setlength{\belowcaptionskip}{-10pt}
\begin{figure*}
\centering
\includegraphics[width=\textwidth]{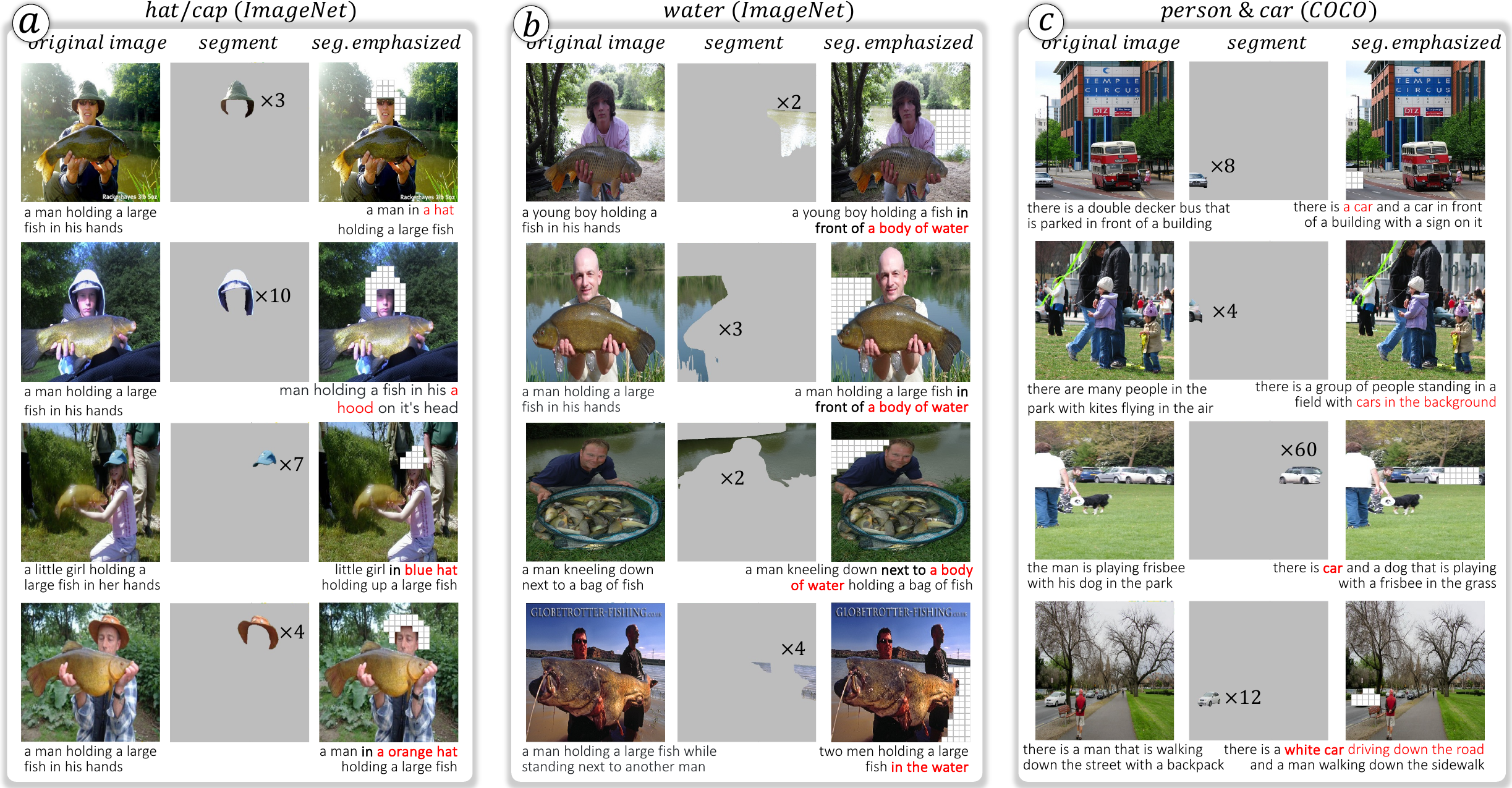}
\vspace{-0.2in}
\caption{ImageNet \texttt{tench} class images with the (a) \texttt{hat} and (b) \texttt{water} features, and (c) the COCO \texttt{person+car} images. The images are randomly picked from the clusters of segments with the respective features. The images' original captions and updated captions are shown below the images.}
\label{fig:caption}
\end{figure*}
\setlength{\belowcaptionskip}{0pt}

Second, users can more explicitly specify what image features the model should look at. In Fig.~\ref{fig:caption_case}a, after updating the prompt to ``\texttt{the man is wearing}'', the word \texttt{headphones} appears, responding to \texttt{wearing} in our prompt. However, the feature of interest is the hat. Therefore, we select the hat patches and increase their weight to 1.5. From the result shown in Fig.~\ref{fig:caption_case}b, we can see \texttt{white hat} successfully appear in the new caption. Further increasing the weight to 2 will make the model even more focus on the \texttt{white hat} region and ignore the \texttt{red shirt}. Note that the weight value for different images could be different and we found the proper value through trial-and-error explorations. Our interface instantly generates new captions while users emphasizing/de-emphasizing different patches, providing immediate feedback to accelerate the trial-and-error explorations. A general heuristic for weight tuning is to keep increasing its value until the desired word (or a similar word) appears. However, an overlarge weight will break the semantics of the caption or make the model repeat certain words. In these cases, we may fail to generate captions with desired words, as the model cannot recognize the corresponding features. 

Fig.~\ref{fig:caption} presents more randomly selected caption steering examples for (a) \texttt{hat}, (b) \texttt{water}, and (c) \texttt{car} features, respectively. For each set of examples, the left, middle, and right columns show the original images and captions, the emphasized segments and applied weights, the updated images and new captions, respectively. In most cases, the weight is between 2 and 10. However, there are also cases where we need to use a very large weight, e.g., 60 in the third example of Fig.~\ref{fig:caption}c.

\subsection{Domain Experts' Feedback}
\label{sec:feedback}

After walking the five domain experts ($E_1{\sim}E_5$) through various cases and ensuring that they have a comprehensive understanding of our solution, we gather feedback from them regarding the utility of our system, pros-and-cons of our approach, and further improvement suggestions.

In general, all experts expressed positive views on our idea and the developed system. 
For image exploration, $E_1$ liked the function of flexibly pinning individual nodes through drag-and-drop, which is very helpful to him when navigating the dynamic node-link diagram. While exploring the case in Fig.~\ref{fig:case_wolf}, he noticed that there is a big node for \texttt{dog} and it is not connected to \texttt{wolf}, indicating that the word \texttt{dog} and \texttt{wolf} never appeared together in the same caption. This reflects that the model either correctly recognized the animal as a \texttt{wolf} or misclassified it as a \texttt{dog}. He regarded this as an inherent limitation of BLIP in differentiating wolves from dogs. 
$E_4$ was particularly intrigued by the idea of using automatically generated captions to succinctly summarize large-scale image datasets. He has been working on de-biasing pre-trained word and image embeddings for years. One recurring challenge he faces is quickly confirming the existence of potential biases. Our approach could serve as a valuable starting point before applying his de-biasing algorithms.
$E_3$ and $E_5$ expressed their appreciation for the seamless coordination between the node-link diagram and scatterplot. They further brainstormed using them to hierarchically explore image features. For example, in Fig.~\ref{fig:case_tench_weakness}a, the \texttt{head+body}, \texttt{face+hand}, and \texttt{face} can be considered clusters nested under the \texttt{man} cluster. These clusters are also closely situated in the scatterplot. Unlike our current unsupervised explorations, such a hierarchy would significantly expedite the explorations and provide deeper insights into image features.

For caption coverage and steering, $E_2$ successfully generated the word \texttt{hat} in Fig.~\ref{fig:teaser}c by updating the prompt. Instead of stopping there, he wondered what would happen if further enhancing the \texttt{hat} patches. Then, he selected those patches, increased their weight to 1.5, and generated the caption again. Surprisingly, the new caption replaced \texttt{hat} with \texttt{beanie} (i.e., a hat without brims), which is a more accurate description. Both $E_2$ and $E_5$ were impressed by the associations encoded in cross-attention and appreciated the visualizations that intuitively expose them. $E_5$ liked our idea of computing the coverage for each segment and commented that the segments covered by too many words (i.e., the purple points in Fig.~\ref{fig:case_tench_weakness}a) may also be problematic, as the received attention is too diverse. He suggested using some metrics, like entropy, to quantify the diversity level of the strongly attending words. 

On the other hand, the experts also mentioned some imperfections and desired features. For example, $E_5$ wished the weight adjustments in caption steering could be more automatic and applied to batches of images to minimize human effort. $E_1$ and $E_2$ brainstormed using SAM to assist in selecting patches with desired features, i.e., by clicking on any pixels within the segment to be captioned and having SAM automatically select the related patches. $E_5$ also suggested using large language models to help the prompt-engineering process and generate words that better provoke the model to generate desired words. These suggestions provide promising directions to further extend our work.

%% file: tex/7discussion.tex
\section{Discussion, Limitations, and Future Work}

\textbf{Pre-trained models' quality.} 
Our exploration of image datasets heavily relies on pre-trained language-image models. Consequently, quality issues stemming from these models can impact our exploration. For example, when exploring the \texttt{white\_wolf} images in Fig.~\ref{fig:case_wolf}, there is a big node for \texttt{dog}, indicating the pre-trained BLIP occasionally misclassified the \texttt{wolf} as \texttt{dog} in some images and generated inaccurate captions. This issue has also been recognized by the experts. However, as the availability of high-quality pre-trained models continues to grow, we anticipate that this problem will become less severe in the future.
Additionally, although we focus primarily on BLIP in this work, our system can be effortlessly applied to other image captioning models, as long as they use cross-attention for image-text association, which currently is the most prevailing approach in multi-modal learning.

\textbf{The node-link diagram.} 
Having explored multiple image datasets with our system, we gradually noticed two insufficiencies in our \graph{} and are working on some solutions to improve them. First, despite the force-directed layout, we are considering incorporating word similarity into the nodes' layout. For example, the words \texttt{green}, \texttt{grass}, and \texttt{tree} in Fig.~\ref{fig:case_wolf}b should be automatically positioned closer to help the exploration (now they are manually pinned together through drag-and-drop). The similarity between words can be measured by the distance between their word2vec embeddings. This will be particularly valuable when navigating larger node-link diagrams.
Second, we are also considering implementing a node filter based on the part-of-speech of the corresponding words. For example, the word \texttt{large} in Fig.~\ref{fig:teaser}a and \texttt{white} in Fig.~\ref{fig:case_wolf} are adjectives describing the respective objects. Their substantial size may divert users' attention from analyzing the co-occurred objects, which are typically nouns. Existing packages, like NLTK~\cite{bird2009natural}, can help to accomplish this function easily.

\textbf{Caption steering.} 
We have also identified two limitations in our current caption steering approach. First, we observed that the generated captions can be highly sensitive to the guiding prompt in some cases. For example, in Fig.~\ref{fig:caption_case}a, the word \texttt{headphones} would not appear if simply changing the prompt from ``\texttt{the man is wearing}'' to ``\texttt{the person is wearing}''. Second, we currently can only derive the proper weight for user-interested image patches through a trial-and-error process. The consequences of both limitations require users to iteratively play with our system to generate desired captions. 
On a positive note, the instantly generated captions offer valuable feedback to help users adjust the prompt/weight, which mitigate the limitations to some extent.
In the future, we plan to work with the domain experts towards more automatic prompt engineering and weight searching algorithms. These efforts will reduce the level of required user interactions, making our system more user-friendly and efficient.

%% file: tex/8conclusion.tex
\section{Conclusion}
In this paper, we introduce a visual analytics system to explore large-scale image datasets through the captioning power of a pre-trained language-image model, i.e., BLIP. Our system is also capable of evaluating the generated captions by revealing their coverage over different image features. Focusing on the insufficiently covered features, we provide an interface to steer the caption generation process. Through multiple cases conducted on two large-scale image datasets with multiple domain experts, we validated the effectiveness of our system.

%% file: tex/A_image_exploration.tex
\newpage
\section{More Results on Image Data Exploration}

This section delves into additional examples showcasing the utilization of our \graph{} in the exploration of ImageNet~\cite{imagenet} and COCO~\cite{coco} datasets. These cases serve to underscore the versatility of our system in effectively navigating diverse image classes.

\subsection{Exploring Different Species in ImageNet}

\setlength{\belowcaptionskip}{-5pt}
\begin{figure}[tbh]
\centering
\includegraphics[width=\columnwidth]{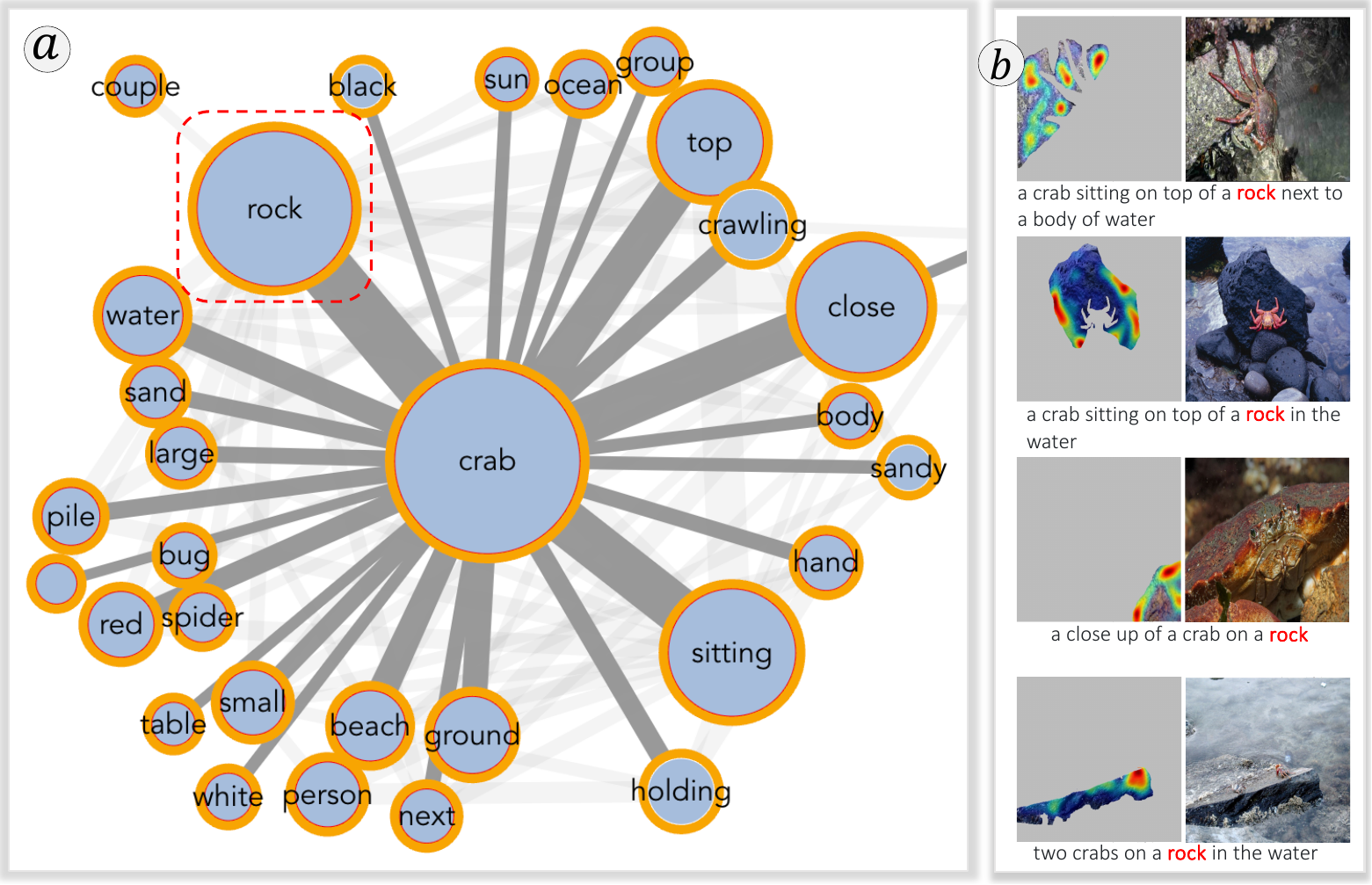}
\vspace{-0.2in}
\caption{(a) For the captions of \texttt{cancer\_irroratus} images, \texttt{rock} and \texttt{crab} are the dominant and most co-occurring features, reflecting the fact that this type of crab prefers a rocky habitat. (b) Some example segments responding strongly to the word \texttt{rock}.}
\label{fig:case_crab}
\end{figure}
\setlength{\belowcaptionskip}{0pt}


While exploring the \texttt{cancer\_irroratus} (Atlantic rock crab) images, we observed that our \graph{} adeptly encapsulates the natural habitat of this crab species. Notably, in the overview depicted in Fig.~\ref{fig:case_crab}a, the word \texttt{rock} immediately emerges as the prominent one, with the highest co-occurrence with the word \texttt{crab}. Clicking on the node for \texttt{rock} grants us access to image segments that are strongly associated with this word. Several illustrative segments are presented in Fig.~\ref{fig:case_crab}b, alongside the corresponding source images and their respective captions.
From the captions, it's noteworthy that the word \texttt{rock} is identified independently rather than as part of the phrase ``rock crab''. This suggests that the captioning model recognizes crabs from the images, but it does not know that they are rock crabs (i.e., the model is not capable of differentiating different types of crabs). The segments that exhibit a strong response to the word \texttt{rock} correspond to the rocky terrains frequently inhabited by \texttt{cancer\_irroratus}, reflecting the typical habitat of this crab species.



\setlength{\belowcaptionskip}{-15pt}
\begin{figure}[tbh]
\centering
\includegraphics[width=.88\columnwidth]{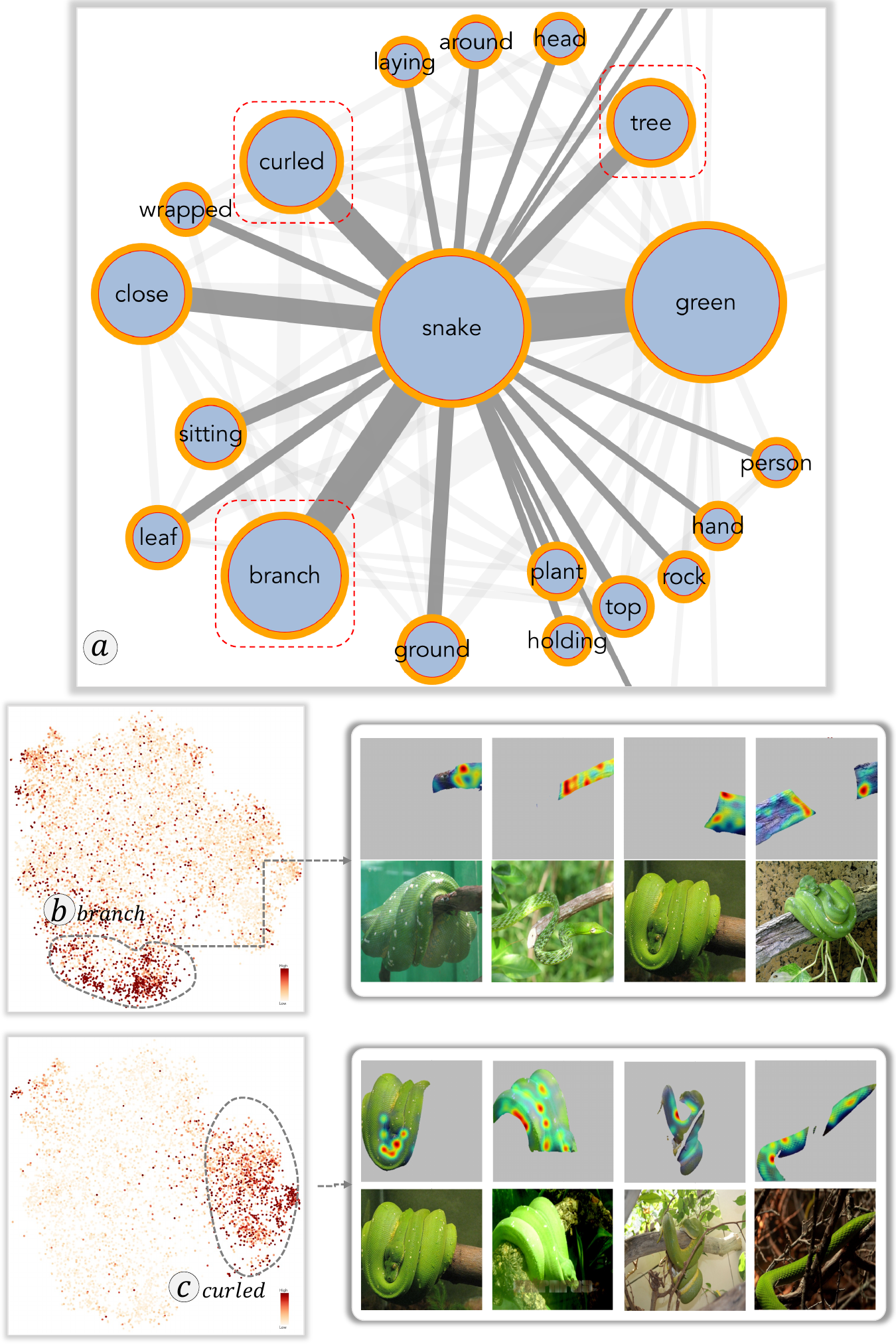}
\vspace{-0.1in}
\caption{(a) The dominant words in the captions for \texttt{green\_snake} images are \texttt{green}, \texttt{branch}, \texttt{tree}, and \texttt{curled}. These words and their co-occurrence relationships depict the living habits and main postures of this type of snake. (b) and (c) show some example segments that strongly respond to the word \texttt{branch} and \texttt{curled}, respectively.}
\label{fig:case_snake}
\end{figure}
\setlength{\belowcaptionskip}{0pt}

Next, by scrutinizing \texttt{green\_snake} images, our system unveils the prevalent living habits of this species. In the \graph{} view for \texttt{green\_snake} (Fig.~\ref{fig:case_snake}a), we initially observed that the word \texttt{green} most frequently co-occurs with \texttt{snake}, indicating the snakes' color. Following \texttt{green}, the three words that emerge with notably high co-occurring frequencies are \texttt{branch}, \texttt{curled}, and \texttt{tree}. This suggests that \texttt{green\_snakes} are commonly found in proximity to a \texttt{tree\_branch} and often exhibit a \texttt{curled} posture. A more in-depth examination of the image segments that strongly respond to the words \texttt{branch} and \texttt{curled} (Fig.~\ref{fig:case_snake}b and \ref{fig:case_snake}c) substantiates the prevalence of this distinctive posture among \texttt{green\_snakes}.

\setlength{\belowcaptionskip}{-10pt}
\begin{figure*}[!tbh]
\centering
\includegraphics[width=\textwidth]{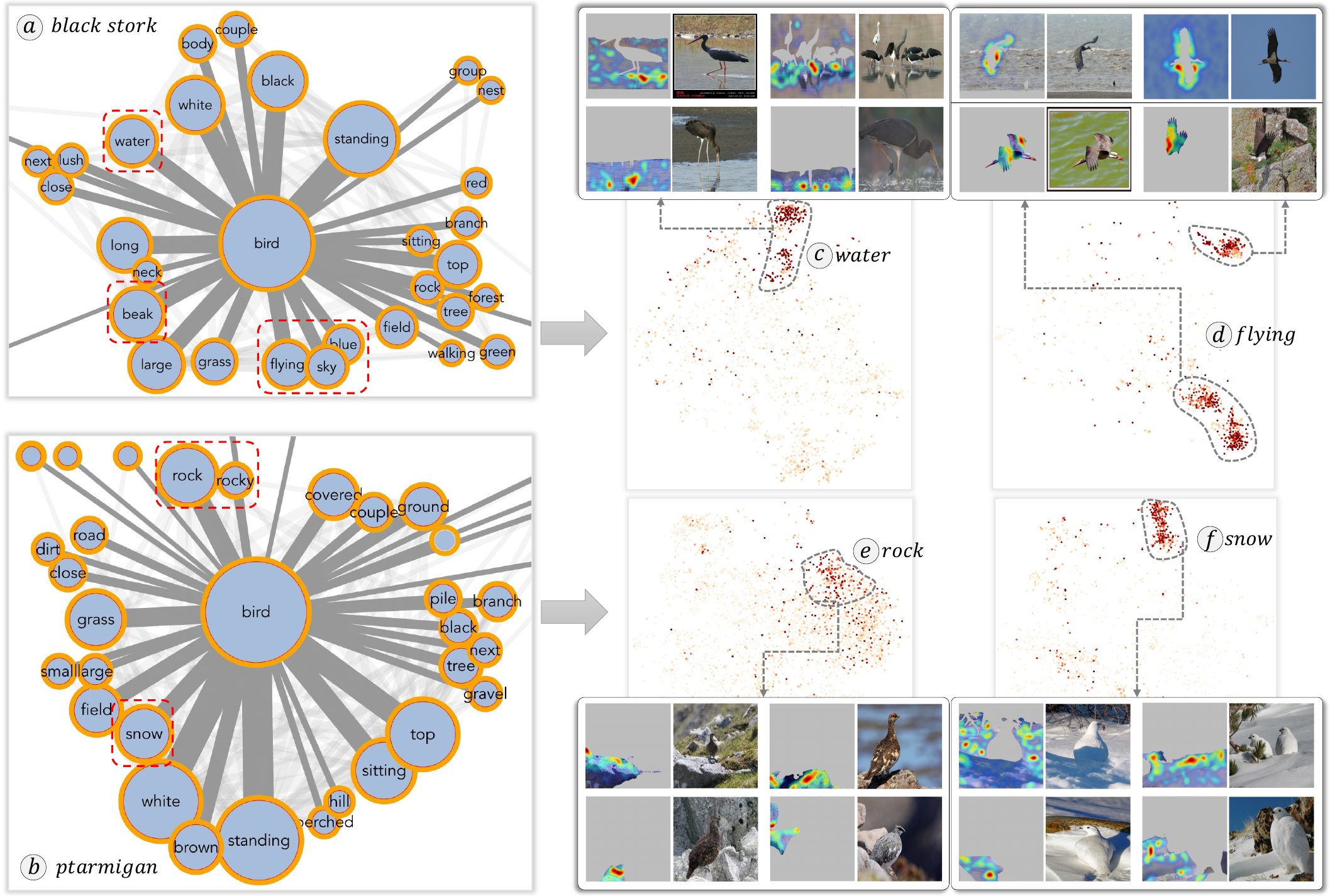}
\vspace{-0.1in}
\caption{Using our system to compare two similar species of birds: \texttt{black\_stork} (a, c, d) and \texttt{ptarmigan} (b, e, f). Our visualizations could reveal the difference of the two birds in their living environment and behavior patterns.}
\label{fig:case_bird}
\end{figure*}
\setlength{\belowcaptionskip}{0pt}

We have also investigated the images of two similar species of birds, \texttt{black\_stork} and \texttt{ptarmigan}, and compared their subtle differences using our system. From the exploration, we have several observations. 
\textit{\textbf{First}}, as we have repeatedly noticed (in the \texttt{tench} and \texttt{crab} cases), the pre-trained BLIP is not able to differentiate different subtypes of birds and use the word \texttt{bird} to describe both species (Fig.~\ref{fig:case_bird}). 
\textit{\textbf{Second}}, \texttt{black\_storks} have long beaks and inhabit mostly around river or marshy areas, whereas \texttt{ptarmigans} have much shorter beaks and inhabit mostly on the highlands. These differences can be easily observed through the respective \graph{}. As shown in Fig.~\ref{fig:case_bird}, the most frequently co-occurred words for \texttt{black\_stork} images are \texttt{water} and \texttt{beak}, while the most frequently co-occurred words for \texttt{ptarmigan} are \texttt{rock} and \texttt{snow}. 
\textit{\textbf{Third}}, the words \texttt{flying}, \texttt{blue} and \texttt{sky} also appear frequently for \texttt{black\_stork}, but are not observed from the node-link diagram for \texttt{ptarmigan}. This indicates another difference between the two species: \texttt{black\_storks} are more often found flying (Fig.~\ref{fig:case_bird}d), while \texttt{ptarmigans} prefer staying on the ground. 
\textit{\textbf{Lastly}}, by exploring the segments extracted from \texttt{ptarmigan} images that respond strongly to the words \texttt{rock} and \texttt{snow}, we also noticed that the plumage color of \texttt{ptarmigan} changes from brown to white in different seasons (i.e., seasonal camouflage) to better hide themselves into the environment. As shown in Fig.~\ref{fig:case_bird}e and \ref{fig:case_bird}f, the \texttt{ptarmigan} with \texttt{rock} and \texttt{snow} background are in brown and white colors, respectively.

\subsection{Potential Biases with Human-Related Objects in COCO}

Next, we focus on several human-related objects from COCO images to analyze the co-occurring features within each object. 

\setlength{\belowcaptionskip}{-10pt}
\begin{figure*}[!tbh]
\centering
\includegraphics[width=\textwidth]{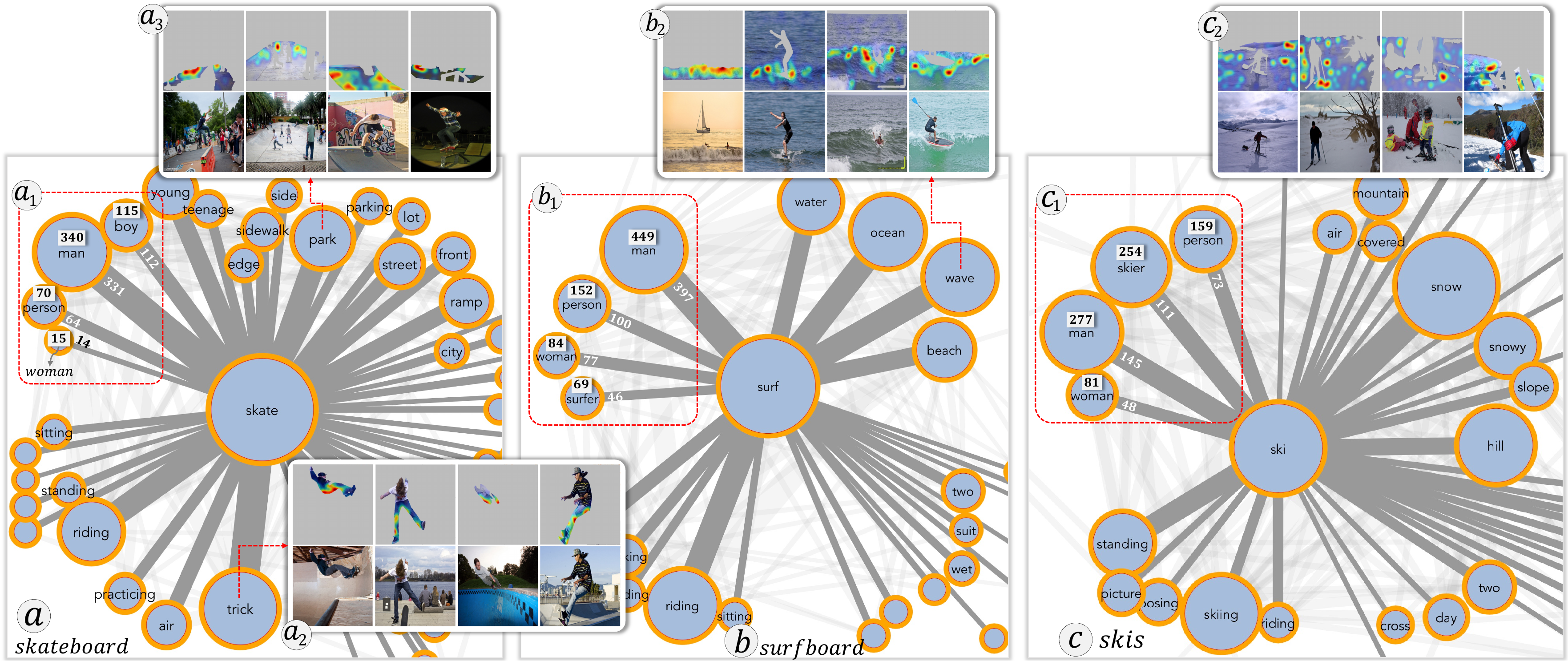}
\vspace{-0.2in}
\caption{Exploring the feature co-occurrences of several sport-related objects from COCO: (a) \texttt{skateboard}, (b) \texttt{surfboard}, (c) \texttt{skis}. The randomly selected segments and their residing images demonstrate examples of the highly co-occurred features.}
\label{fig:case_sports}
\end{figure*}
\setlength{\belowcaptionskip}{0pt}

\setlength{\belowcaptionskip}{-10pt}
\begin{figure}[!tbh]
\centering
\includegraphics[width=\columnwidth]{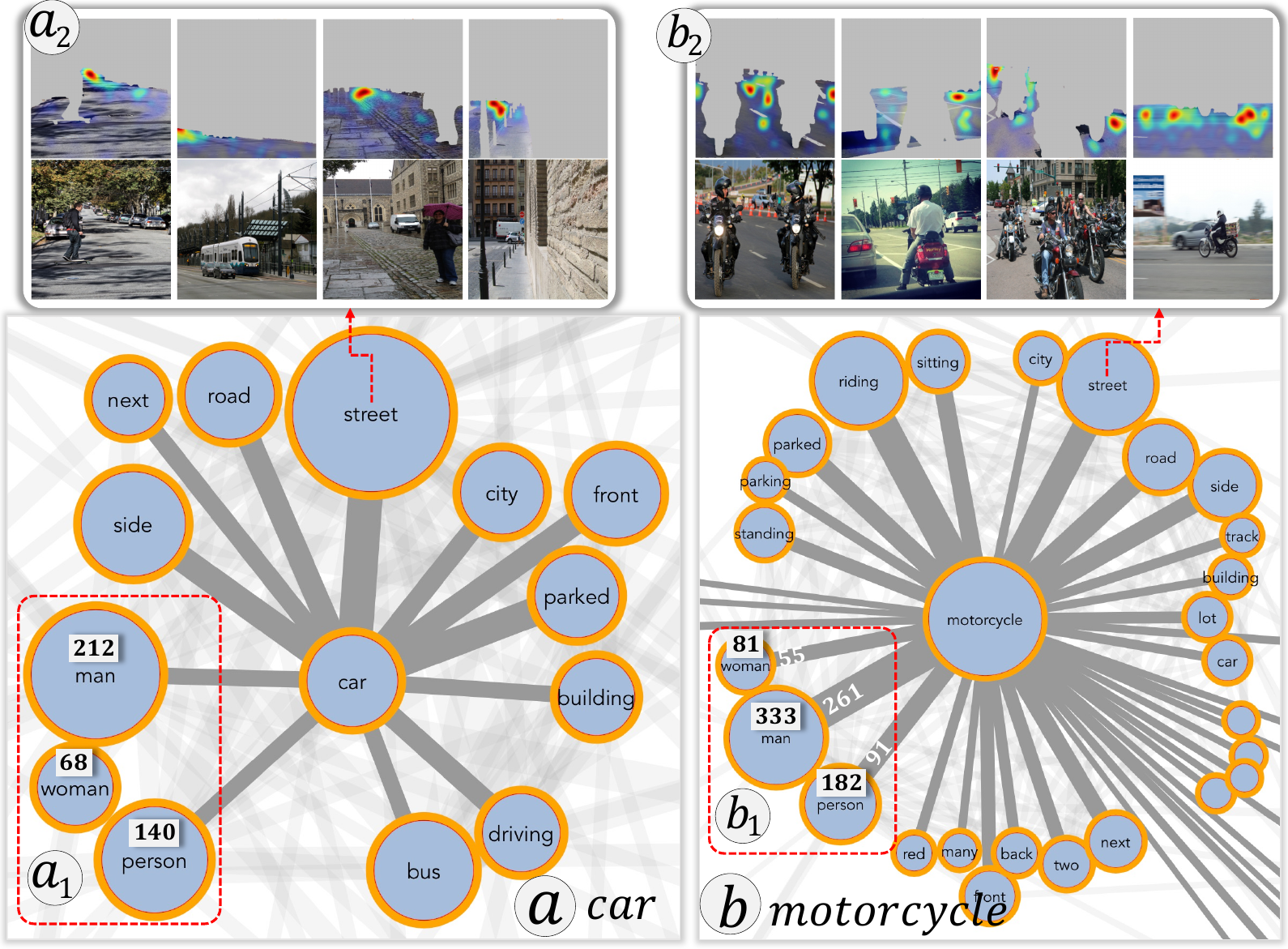}
\vspace{-0.2in}
\caption{Exploring the feature co-occurrences of several vehicle objects from COCO: (a) \texttt{car}, (b) \texttt{motorcycle}.}
\label{fig:case_vehicles}
\end{figure}
\setlength{\belowcaptionskip}{0pt}

We first attempted to identify potential gender biases in images of various sport equipment, i.e., \texttt{skateboard}, \texttt{surfboard}, and \texttt{skis} (Fig.~\ref{fig:case_sports}),  and vehicles, i.e., \texttt{car} and \texttt{motorcycle} (Fig.~\ref{fig:case_vehicles}). 
From the \graph{} view, we observed a common pattern among these objects where the frequencies of \texttt{woman} and \texttt{person} were significantly lower than that of \texttt{man} (see Fig.~\ref{fig:case_sports}-a1, b1, c1, and Fig.~\ref{fig:case_vehicles}-a1, b1).
Notably, not only did \texttt{man} have the highest occurrence in these images (see the corresponding nodes' size), but it also surpassed \texttt{woman} and \texttt{person} in terms of co-occurrence frequencies with the sport or vehicle objects in focus (see the corresponding links' width). 
One extreme example is found within the \texttt{car} images, where there was no link between \texttt{car} and \texttt{woman} (Fig.~\ref{fig:case_vehicles}-a1). The exact frequency values are annotated above each node and link, providing a basis for these comparisons.
These imbalanced gender occurrences in the images highlight potential imperfections in image sampling, which may lead an ML model to exhibit biased behaviors in gender-related tasks. 


Meanwhile, we have also explored other contextual features that prominently co-occurred with the target objects. For example, among the \texttt{skateboard} images, the most commonly co-occurring words include \texttt{trick} and \texttt{park}. 
These words are associated with the actions of skateboarders (e.g., performing a \texttt{trick}) and the skate-park environment (Fig.~\ref{fig:case_sports}-a2, a3), shedding light on the dominant human activities and settings related to \texttt{skateboard}. 
Similar dominance of environmental features has also been identified in \texttt{surfboard} images (e.g., \texttt{wave} in Fig.~\ref{fig:case_sports}-b2) and \texttt{skis} images (e.g., \texttt{snow} in Fig.~\ref{fig:case_sports}-c2). For the vehicle images, like \texttt{car} and \texttt{motorcycle}, the most frequent co-occurring feature is \texttt{street} (Fig.~\ref{fig:case_vehicles}-a2 and b2), which indicates that these vehicles are mostly depicted in urban street settings.

    

%% file: tex/B_caption_steering.tex
\section{More Results of Caption Coverage and Steering}



\begin{figure*}[!tbh]
\centering
\includegraphics[width=\textwidth]{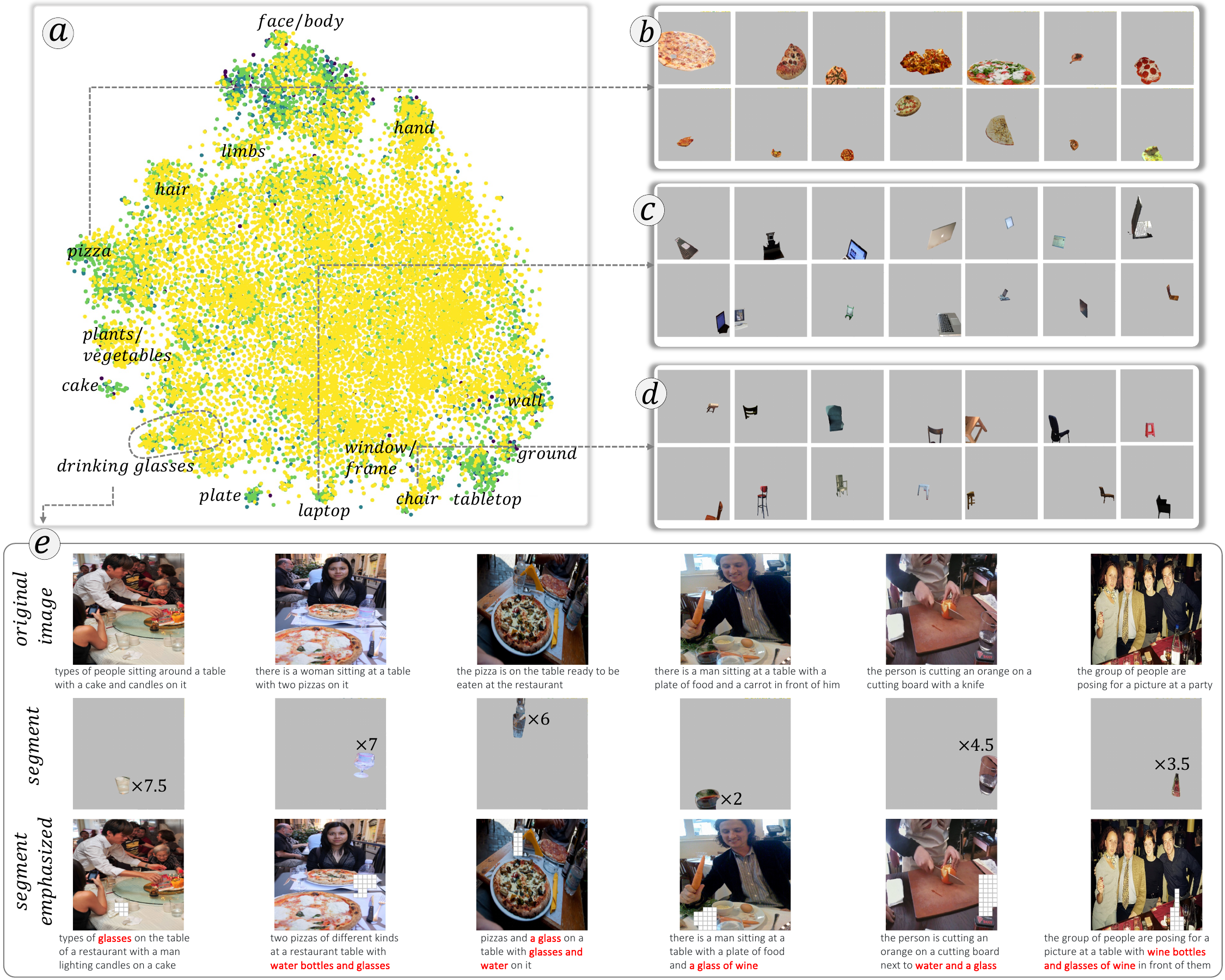}
\vspace{-0.25in}
\caption{(a) The clusters of image segments from the \texttt{dining\_table} images. (b, c, d) Lasso-selection results of the \texttt{pizza}, \texttt{laptop}, and \texttt{chair} segments. (e) Steering the image captioning model to generate descriptions for \texttt{drinking\_glasses} by emphasizing the corresponding segments.}
\label{fig:case_dining_table}
\end{figure*}

\setlength{\belowcaptionskip}{-10pt}
\begin{figure*}[!tbh]
\centering
\includegraphics[width=\textwidth]{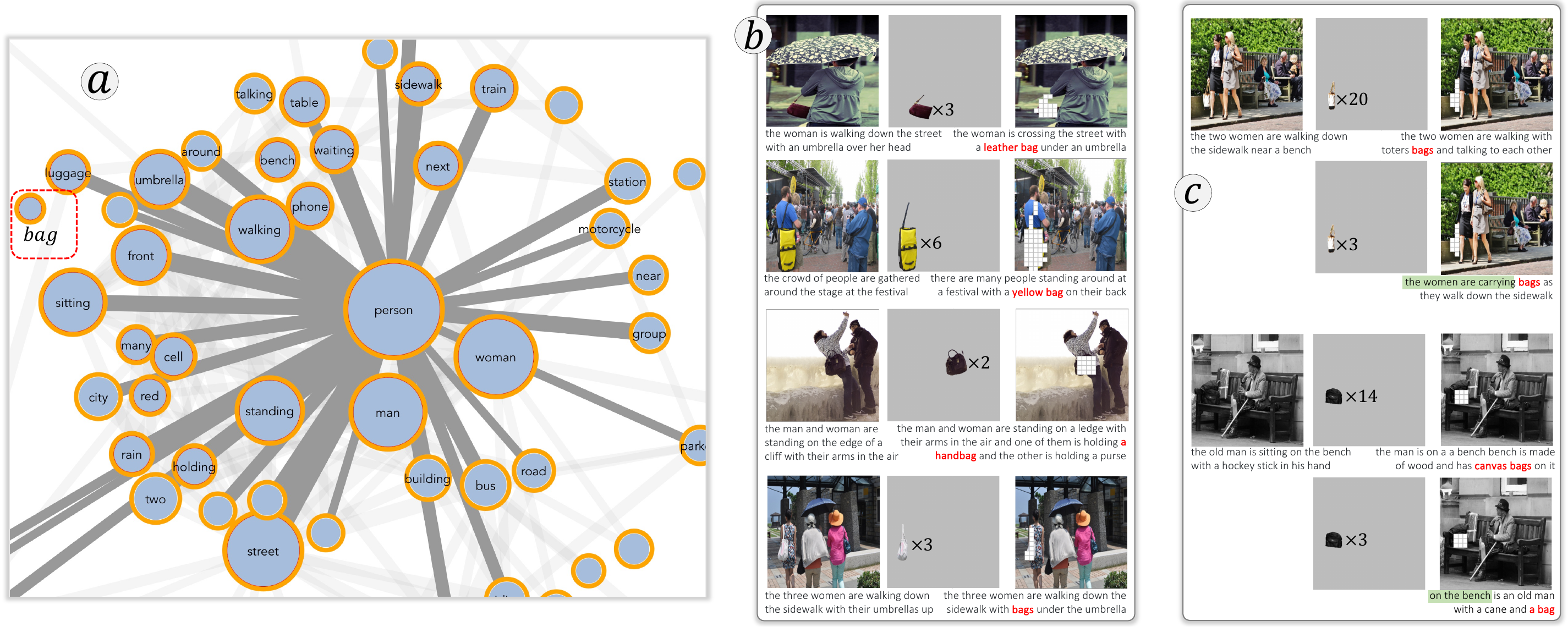}
\vspace{-0.2in}
\caption{(a) The \graph{} view of \texttt{handbag} images. (b) Steering the image captioning model by adjusting segments' weight only. (c) Steering the image captioning model by adjusting the segment's weight and modifying the starting prompts.}
\label{fig:case_bag}
\end{figure*}
\setlength{\belowcaptionskip}{0pt}


This section includes more explorations with our system on caption coverage and caption steering, in addition to the \texttt{tench} and \texttt{person+car} cases that we have shown in Sec. 6.2 and Sec. 6.3 of the paper.

First, we examined the COCO images that feature both \texttt{person} and \texttt{dining\_table}. Several clusters with common features can be identified through the \scatterplot{} view (Fig.~\ref{fig:case_dining_table}a). These clusters include human-related features, such as \texttt{face}, \texttt{body}, and \texttt{hair}. They also include features associated with dining tables, such as \texttt{pizza}, \texttt{laptop}, and \texttt{chair} (shown in Fig.~\ref{fig:case_dining_table}b-d). Among these clusters, we noticed that many image segments within the \texttt{drinking\_glasses} cluster are not sufficiently covered by the captions, but this feature is an essential object on a \texttt{dining\_table}. Therefore, we target to steer the image captioning model to generate new captions describing this feature. 
Several caption steering results are shown in Fig.~\ref{fig:case_dining_table}e. With the guidance of the \texttt{drinking\_glasses} segments (the middle row), we are able to select and scale up the corresponding image patches and update the captions to generate the desired feature descriptions (the bottom row).

Next, we move on to explore another set of COCO images featuring both \texttt{handbag} and \texttt{person} objects. To our surprise, we discovered that our target feature, i.e., the \texttt{bag}, is rarely mentioned in the image captions and seldom co-occurs with other objects from the \graph{} view, as highlighted in Fig.~\ref{fig:case_bag}a.
As we delved deeper into these images and scrutinized the \texttt{handbag} segments, we realized that the \texttt{handbags} depicted in most of them are smaller in scale compared to other objects, such as \texttt{person}, \texttt{bench}, and \texttt{umbrella}.
As a result, these smaller \texttt{handbags} are often unnoticed during the caption generation process.
In Fig.~\ref{fig:case_bag}b, we present four example images and the corresponding segments, where we lay more emphasis on the \texttt{handbag} patches and regenerate the captions. 
Utilizing solely the segment scaling technique, we successfully ensured the inclusion of the word \texttt{bag} in all four images' new captions.
The two cases in Fig.~\ref{fig:case_bag}c, however, require much larger scaling weights. With additional guidance from the starting prompts, we are able to generate desired captions with much smaller segment scaling weights.



%% file: tex/C_batch_processing.tex
\begin{figure*}[!tbh]
\includegraphics[width=\textwidth]{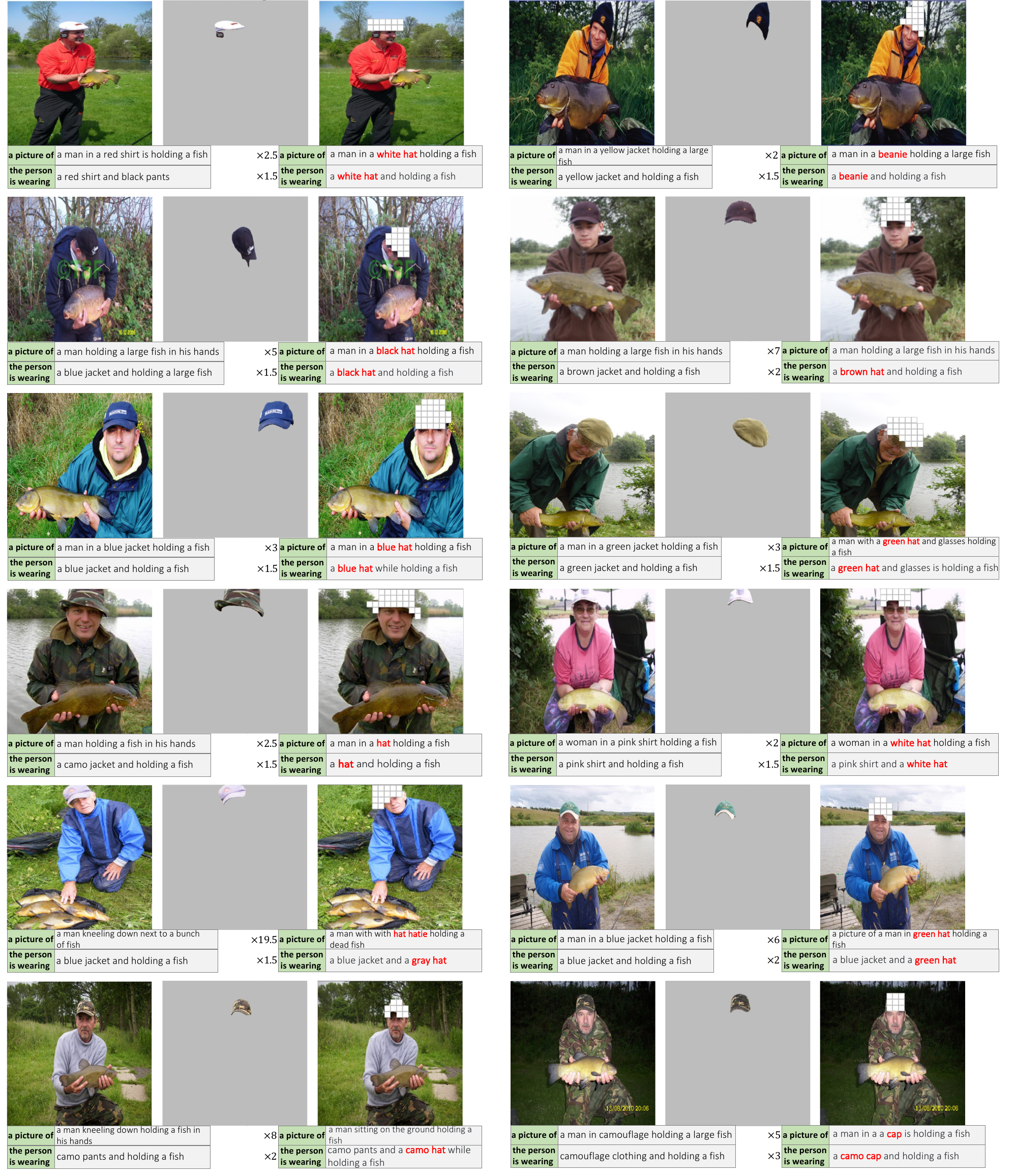}
\caption{The 12 images that failed to generate descriptions for \texttt{hat} features when applying prompt steering only. Explicitly enhancing the \texttt{hat} segments or using both prompt-engineering and segment enhancement successfully guide the model to generate descriptions for \texttt{hat} in the new captions.}
\label{fig:hat_patch_steer}
\end{figure*}

\begin{figure}[!tbh]
\includegraphics[width=\columnwidth]{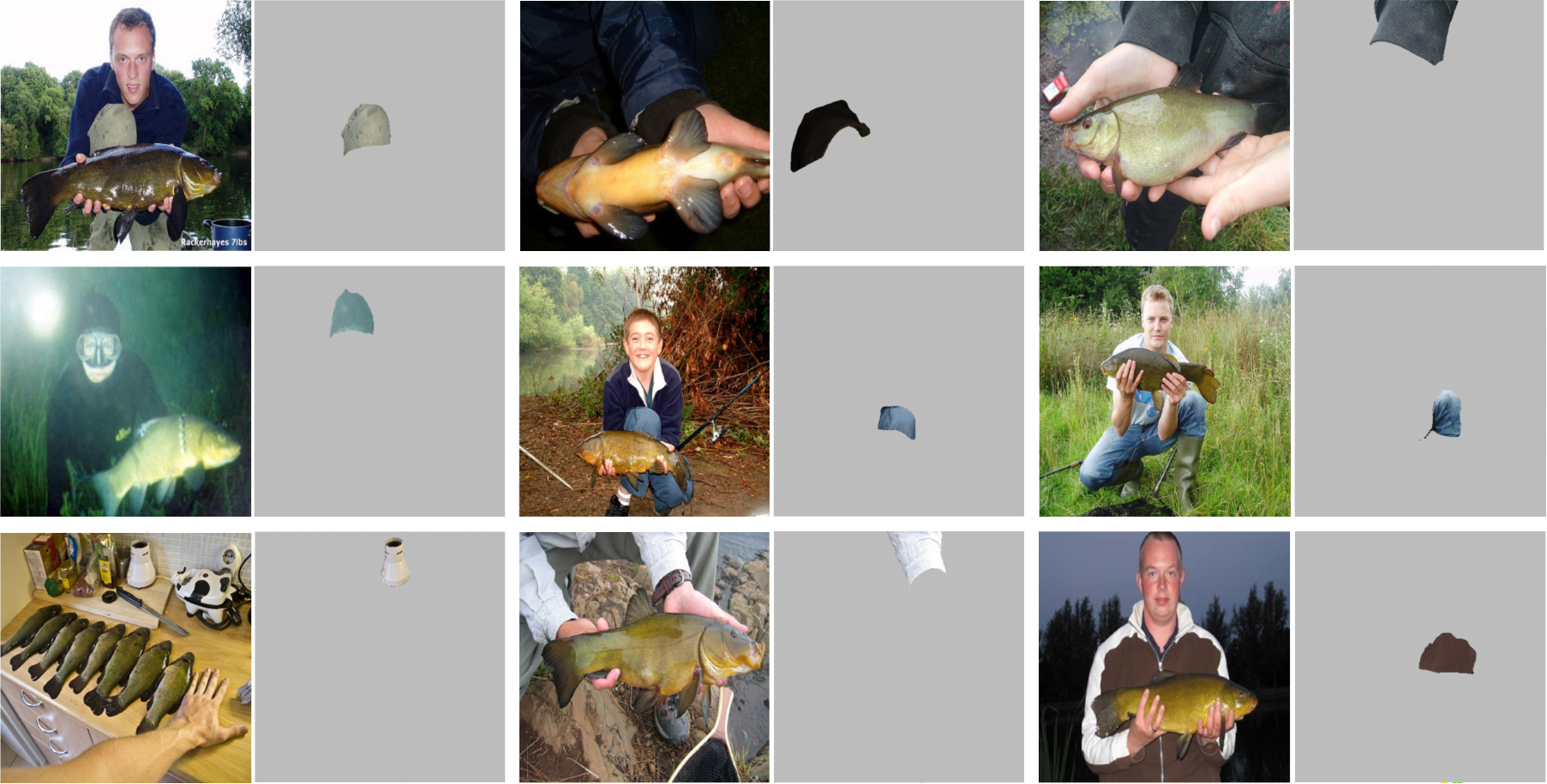}
\caption{The extracted segments from these 9 images are very similar to hat features. However, they are actually not hat and we cannot steer the model to generate captions involving the word \texttt{hat}.}
\label{fig:images_no_hat}
\end{figure}

\section{Batch Processing Results of Caption Steering}

In Sec. 6.3 of the paper, we explained one case in which we applied a new prompt to a batch of images to generate new captions. Here, we provide more details to support that case. Specifically, we lasso selected a cluster of image segments representing the hat features in the \texttt{tench} images. After removing duplicated images and the images with \texttt{hat} in their captions, 214 images were identified. These images contain hat-like segments but their captions lack descriptions for the hat features.
We changed the default prompt from ``\texttt{a picture of}'' to ``\texttt{the person is wearing}'' for these $214$ images and regenerated their captions. Among the $214$ new captions, $193$ contain words describing the hat features (e.g., \texttt{hat}/\texttt{hood}/\texttt{beanie}), resulting in a success rate of 90.19\%. 
A detailed summary of the caption steering results, along with the information of images, can be found in our supplementary Excel spreadsheet.

Next, we inspected the $21$ ($214{-}193$) failed cases. 
Among them, $12$ images indeed have the hat feature but the updated prompt cannot help to generate descriptions for it in the captions. 
For each of the $12$ images, we then applied the caption steering in two ways: (1) scaling up the hat patches with the original prompt ``\texttt{a picture of}'', and (2) simultaneously scaling up the hat patches and using the new prompt ``\texttt{the person is wearing}''. The experiment results are presented in Fig.~\ref{fig:hat_patch_steer}. 
For the first method of scaling up the patches' weight only, the model still failed on some images. However, it was able to generate descriptions for the hat features in all 12 images' captions, when using the second method, i.e., adjusting both the prompt and the patches' weight. 
For the remaining $9$ ($21{-}12$) images, we found that they actually do not contain hat features, but the SAM extracted features from them are very similar to a hat, as shown in Fig.~\ref{fig:images_no_hat}. This detailed exploration explains why the model fails in some cases and further validates the good performance of our caption steering methods.


%

%% file: tex/D_gradcam_evaluation.tex
\section{Grad-CAM Evaluation and Computation Details}

In Sec. 5.2 of the paper, we explained how Grad-CAM is computed (Fig. 5 of the paper) and used to disclose the association between image segments and caption words. This section provides more details to support the evaluation and computation of Grad-CAM. 

\subsection{Cross-Attention Evaluation}
\label{sec:gradcam_eval}

There are two groups of cross-attentions from the BLIP model (see Fig. 2 of the paper). One is from the LM part and the other is from the ITM part. Each group has 144 copies of the cross-attention matrix, since each part is a Transformer with 12 layers and 12 heads. 
Aggregating the cross-attentions from the 12 heads of the same layer is commonly adopted in previous works~\cite{li2021align} and we employed the same operation. Therefore, we only need to compare the aggregated attentions from 12 layers. 
Additionally, we also want to show that using Grad-CAM is better than using cross-attention (CA) only. That is, the matrix $C$ in Fig. 5 of the paper is better than matrix $A$ in reflecting the association between image features and caption words. These two perspectives generate four combinations: \texttt{LM\_CA}, \texttt{ITM\_CA}, \texttt{LM\_Grad-CAM}, and \texttt{ITM\_Grad-CAM}. 

\begin{figure}[!tbh]
\centering
\includegraphics[width=\columnwidth]{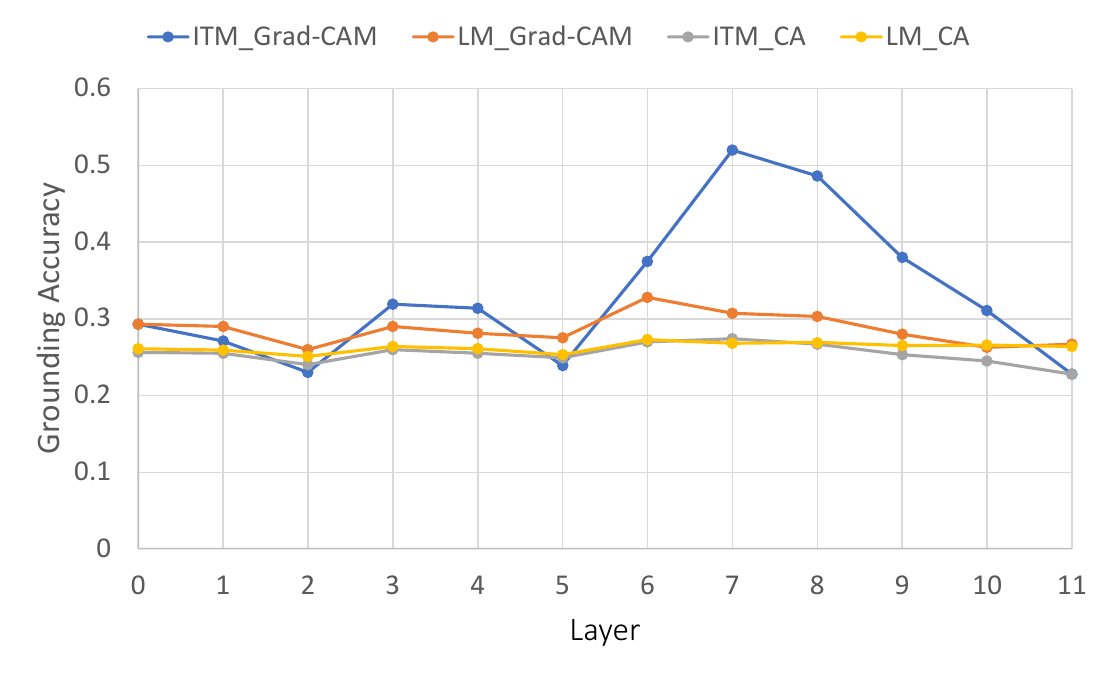}
\caption{Layer-wise evaluation of 4 different association computations. As the legend shows, Grad-CAM means using the element-wise multiplication of gradients and cross-attentions, and CA means using the cross-attentions only. ITM and LM represent using the corresponding parts of the BLIP model. Among all combinations, the best performance is from the Grad-CAM of layer 7 from ITM.}
\label{fig:layer_gradcam_eval}
\end{figure}

Fig.~\ref{fig:layer_gradcam_eval} shows the performance of the four methods. The horizontal axis denotes the 12 layers, the vertical axis presents the grounding accuracy (the higher the better). The grounding accuracy is computed by using the CA or Grad-CAM matrix on the RefCOCO+ dataset~\cite{kazemzadeh2014referitgame} and comparing the matrix-encoded association with the ground-truth association.
Specifically, for each image-caption pair from the RefCOCO+ dataset, people have annotated which word is associated with what image features (visual objects). 
Considering this as the ground-truth, we can check if the CA or Grad-CAM matrix learned the correct attention between caption words and image patches. 
Eventually, we report the grounding accuracy, which is the ratio between the image-caption pairs that the captions accurately attend to the correct visual objects and all image-caption pairs.
Based on the results shown in Fig.~\ref{fig:layer_gradcam_eval}, the Grad-CAM computed from the ITM part of BLIP achieves the best performance at layer 7, and the accuracy value is much higher than other methods or layers. Therefore, we chose to use the Grad-CAM extracted from layer 7 of ITM in our visual analysis by default. However, as we have explained in the paper, users can also flexibly switch to other methods/layers, using the widgets from the headers of the \scatterplot{} view.

\begin{figure}[!tbh]
\centering
\includegraphics[width=\columnwidth]{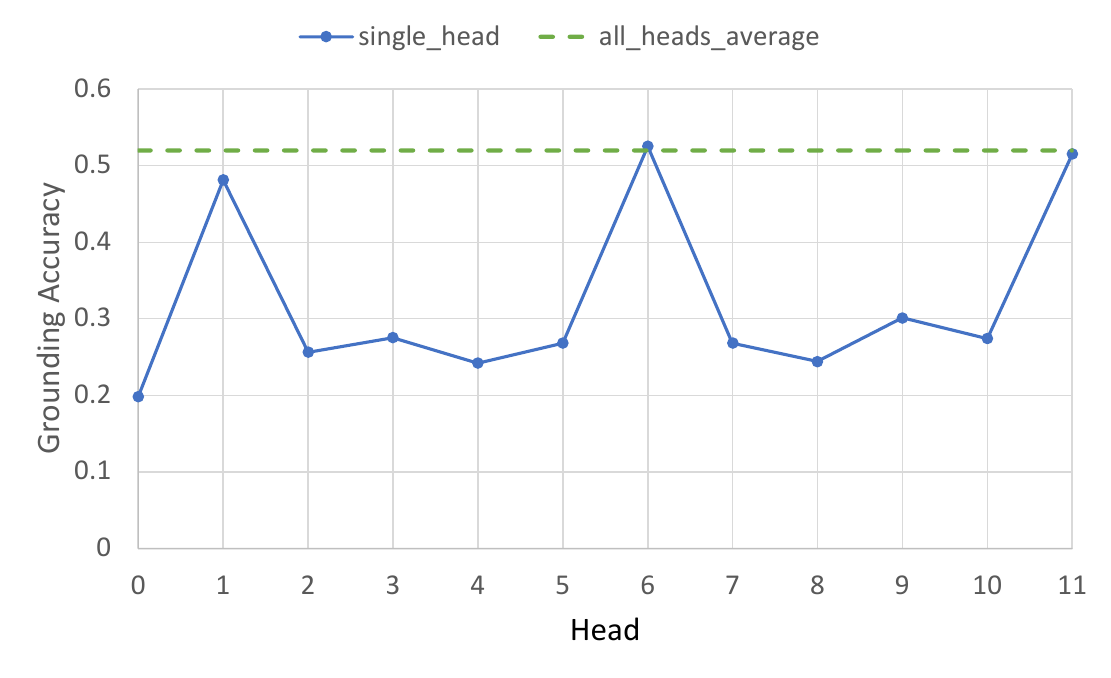}
\caption{Evaluation of individual heads' Grad-CAM within the best-performing layer 7. Three heads (1, 6, and 11) stand out with much higher grounding accuracy than others. The best-performing heads behave similarly to the Grad-CAM averaged over all heads.}
\label{fig:head_gradcam_eval}
\end{figure}

Focusing on the Grad-CAM from layer 7 of ITM, we then drilled down to individual heads of this layer to see their grounding accuracy. As shown in Fig.~\ref{fig:head_gradcam_eval}. Among the 12 heads of layer 7, head 1, head 6, and head 11 achieve better grounding performance than the other ones. The grounding accuracy of Grad-CAM averaged over all heads of this layer is also plotted for reference, which is on par with the best performance of individual heads. This result further supports us in aggregating the 12 heads from individual layers.


\subsection{The Computation Details of Grad-CAM}

\begin{figure}[!tbh]
\centering
\includegraphics[width=\columnwidth]{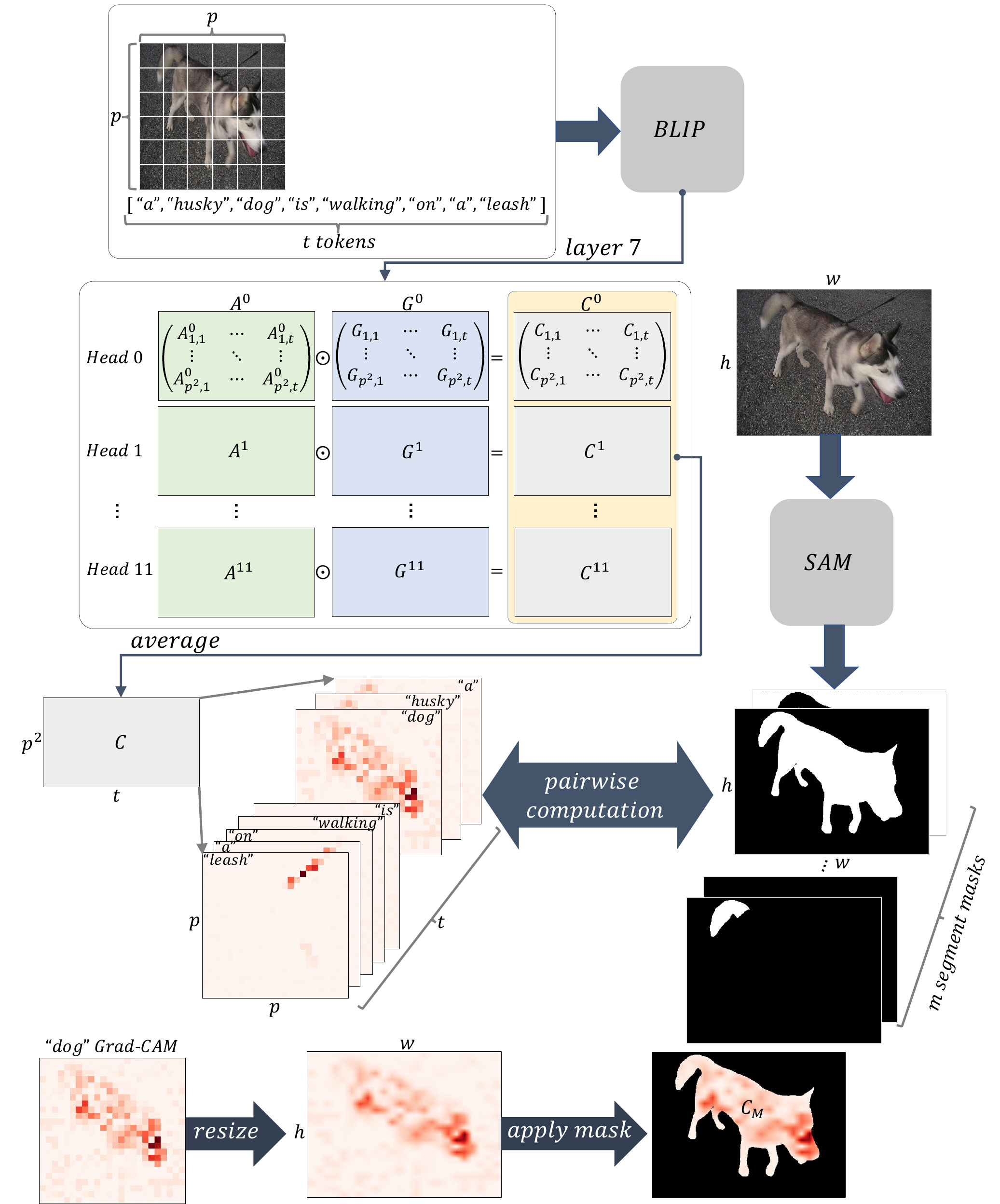}
\caption{Computation details of Grad-CAM aggregations for each word-segment pair.}
\label{fig:gradcam_comp}
\end{figure}

To help people reproduce our experiments, we provide further details on the computation of Grad-CAM to support our explanations in Sec. 5.2 of the paper. The details are illustrated in Fig.~\ref{fig:gradcam_comp} and the computation process can be elaborated with the following steps.

\begin{enumerate}
    \item For each image-caption pair, we decompose the image into $p{\times}p$ patches and the caption into $t$ tokens.  The two parts are then fed into the VIT and ITM parts of BLIP, respectively.
    \item From BLIP, we extract the cross attentions ($A^0$ to $A^{11}$) and gradients ($G^0$ to $G^{11}$) over all heads in the best-performing layer, i.e., layer 7 based on our evaluation in Sec. \ref{sec:gradcam_eval}. Both $A$ and $G$ are with the shape $p^2{\times}t$ (see Fig. 5 of our paper for the notations).
    \item We compute the Grad-CAM of each head using the corresponding $A^i$ and $G^i$, and obtain an aggregated Grad-CAM $C$ averaged over all $C^i$ from individual heads. $C$ is also with the shape of $p^2{\times}t$. 
    \item $C$ is then reshaped into $t$ Grad-CAM maps, each with a shape of $p{\times}p$ and representing the distribution of the Grad-CAM from the corresponding word to all image patches.
    \item In the meantime, we also feed the original $w{\times}h$ image into the SAM model, and generate $m$ segments. Each is represented as a binary mask with the shape of $w{\times}h$.
    \item Using the $p{\times}p$ Grad-CAM maps from Step 4 and the $w{\times}h$ segment masks from Step 5, we aggregate the Grad-CAM for each pair of text token and image segment (word-segment pair) through the following steps:
    \begin{enumerate}
        \item Resize the text token's Grad-CAM map from $p{\times}p$ to $w{\times}h$.
        \item Apply the corresponding segment mask (shape: $w{\times}h$) onto the resized Grad-CAM map (shape: $w{\times}h$), i.e., filter the Grad-CAM map to only contain values within the segment. The filtered map is called $C_M$.
        \item Compute the Grad-CAM score for this segment as $\Sigma C_M / \sqrt{Area(C_M)}$, where $\Sigma C_M$ is the sum of values within the filtered Grad-CAM map, and $Area(C_M)$ is the number of pixels in the segment.
    \end{enumerate}
\end{enumerate}

%% file: main.bbl
\begin{thebibliography}{10}

\bibitem{textblob}
Textblob: Simplified text processing.
\newblock \url{https://textblob.readthedocs.io/}.
\newblock Accessed: 2023-09-19.

\bibitem{aflalo2022vl}
E.~Aflalo, M.~Du, S.-Y. Tseng, Y.~Liu, C.~Wu, N.~Duan, and V.~Lal.
\newblock Vl-interpret: An interactive visualization tool for interpreting vision-language transformers.
\newblock In {\em Proceedings of the IEEE/CVF Conference on Computer Vision and Pattern Recognition}, pp. 21406--21415, 2022.

\bibitem{bahdanau2014neural}
D.~Bahdanau, K.~Cho, and Y.~Bengio.
\newblock Neural machine translation by jointly learning to align and translate.
\newblock {\em arXiv preprint arXiv:1409.0473}, 2014.

\bibitem{bertucci2022dendromap}
D.~Bertucci, M.~M. Hamid, Y.~Anand, A.~Ruangrotsakun, D.~Tabatabai, M.~Perez, and M.~Kahng.
\newblock Dendromap: Visual exploration of large-scale image datasets for machine learning with treemaps.
\newblock {\em IEEE Transactions on Visualization and Computer Graphics (IEEE VIS 2022 Conference)}, 2022.

\bibitem{bilal2017convolutional}
A.~Bilal, A.~Jourabloo, M.~Ye, X.~Liu, and L.~Ren.
\newblock Do convolutional neural networks learn class hierarchy?
\newblock {\em IEEE Trans. Vis. Comput. Graphics}, 24(1):152--162, 2017.

\bibitem{bird2009natural}
S.~Bird, E.~Klein, and E.~Loper.
\newblock {\em Natural language processing with Python: analyzing text with the natural language toolkit}.
\newblock " O'Reilly Media, Inc.", 2009.

\bibitem{cao2020analyzing}
K.~Cao, M.~Liu, H.~Su, J.~Wu, J.~Zhu, and S.~Liu.
\newblock Analyzing the noise robustness of deep neural networks.
\newblock {\em IEEE Transactions on Visualization and Computer Graphics}, 27(7):3289--3304, 2020.

\bibitem{chen2021towards}
C.~Chen, J.~Wu, X.~Wang, S.~Xiang, S.-H. Zhang, Q.~Tang, and S.~Liu.
\newblock Towards better caption supervision for object detection.
\newblock {\em IEEE Transactions on Visualization and Computer Graphics}, 28(4):1941--1954, 2021.

\bibitem{imagenet}
J.~Deng, W.~Dong, R.~Socher, L.-J. Li, K.~Li, and L.~Fei-Fei.
\newblock Imagenet large scale visual recognition challenge.
\newblock {\em International Journal of Computer Vision (IJCV)}, 115(3):211--252, 2015. \href{https://doi.org/10.1007/s11263-015-0816-y}
{doi: {{%
10\hspace{.1pt}\discretionary{.}{%
}{.}\hspace{.4pt}1007\discretionary{/}{%
}{/}s11263\discretionary{%
}{-}{-}015\discretionary{%
}{-}{-}0816\discretionary{%
}{-}{-}y}}}


\bibitem{derose2020attention}
J.~F. DeRose, J.~Wang, and M.~Berger.
\newblock Attention flows: Analyzing and comparing attention mechanisms in language models.
\newblock {\em IEEE Trans. Vis. Comput. Graphics}, 27(2):1160--1170, 2020.

\bibitem{devlin2019bert}
J.~Devlin, M.-W. Chang, K.~Lee, and K.~Toutanova.
\newblock {BERT}: Pre-training of deep bidirectional transformers for language understanding.
\newblock In {\em Proceedings of the 2019 Conference of the North {A}merican Chapter of the Association for Computational Linguistics: Human Language Technologies, Volume 1 (Long and Short Papers)}, pp. 4171--4186, 2019.

\bibitem{dong2020interactive}
Z.~Dong, T.~Wu, S.~Song, and M.~Zhang.
\newblock Interactive attention model explorer for natural language processing tasks with unbalanced data sizes.
\newblock In {\em 2020 IEEE Pacific Visualization Symposium}, pp. 46--50, 2020.

\bibitem{dosovitskiy2020image}
A.~Dosovitskiy, L.~Beyer, A.~Kolesnikov, D.~Weissenborn, X.~Zhai, T.~Unterthiner, M.~Dehghani, M.~Minderer, G.~Heigold, S.~Gelly, J.~Uszkoreit, and N.~Houlsby.
\newblock An image is worth 16x16 words: Transformers for image recognition at scale.
\newblock In {\em 9th International Conference on Learning Representations (ICLR), Austria, May 3-7, 2021}, 2021.

\bibitem{gou2020vatld}
L.~Gou, L.~Zou, N.~Li, M.~Hofmann, A.~K. Shekar, A.~Wendt, and L.~Ren.
\newblock Vatld: A visual analytics system to assess, understand and improve traffic light detection.
\newblock {\em IEEE transactions on visualization and computer graphics}, 27(2):261--271, 2020.

\bibitem{hohman2019s}
F.~Hohman, H.~Park, C.~Robinson, and D.~H.~P. Chau.
\newblock Summit: Scaling deep learning interpretability by visualizing activation and attribution summarizations.
\newblock {\em IEEE Trans. Vis. Comput. Graphics}, 26(1):1096--1106, 2019.

\bibitem{jaunet2021visqa}
T.~Jaunet, C.~Kervadec, R.~Vuillemot, G.~Antipov, M.~Baccouche, and C.~Wolf.
\newblock Visqa: X-raying vision and language reasoning in transformers.
\newblock {\em IEEE Trans. Vis. Comput. Graphics}, 28(1):976--986, 2021.

\bibitem{jia2021towards}
S.~Jia, Z.~Li, N.~Chen, and J.~Zhang.
\newblock Towards visual explainable active learning for zero-shot classification.
\newblock {\em IEEE Transactions on Visualization and Computer Graphics}, 28(1):791--801, 2021.

\bibitem{karpathytsne}
A.~Karpathy.
\newblock t-sne visualization of cnn codes.
\newblock \url{https://cs.stanford.edu/people/karpathy/cnnembed/}.
\newblock Accessed: 2023-09-19.

\bibitem{kaul2021improving}
S.~Kaul, D.~Borland, N.~Cao, and D.~Gotz.
\newblock Improving visualization interpretation using counterfactuals.
\newblock {\em IEEE Transactions on Visualization and Computer Graphics}, 28(1):998--1008, 2021.

\bibitem{kazemzadeh2014referitgame}
S.~Kazemzadeh, V.~Ordonez, M.~Matten, and T.~Berg.
\newblock Referitgame: Referring to objects in photographs of natural scenes.
\newblock In {\em Proceedings of the 2014 conference on empirical methods in natural language processing (EMNLP)}, pp. 787--798, 2014.

\bibitem{khan2021transformers}
S.~Khan, M.~Naseer, M.~Hayat, S.~W. Zamir, F.~S. Khan, and M.~Shah.
\newblock Transformers in vision: A survey.
\newblock {\em ACM Computing Surveys}, 2021.

\bibitem{kirillov2023segment}
A.~Kirillov, E.~Mintun, N.~Ravi, H.~Mao, C.~Rolland, L.~Gustafson, T.~Xiao, S.~Whitehead, A.~C. Berg, W.-Y. Lo, et~al.
\newblock Segment anything.
\newblock {\em arXiv preprint arXiv:2304.02643}, 2023.

\bibitem{li2023lavis}
D.~Li, J.~Li, H.~Le, G.~Wang, S.~Savarese, and S.~C. Hoi.
\newblock {LAVIS}: A one-stop library for language-vision intelligence.
\newblock In {\em Proceedings of the 61st Annual Meeting of the Association for Computational Linguistics (Volume 3: System Demonstrations)}, pp. 31--41. Association for Computational Linguistics, Toronto, Canada, July 2023.

\bibitem{li2022blip}
J.~Li, D.~Li, C.~Xiong, and S.~Hoi.
\newblock Blip: Bootstrapping language-image pre-training for unified vision-language understanding and generation.
\newblock In {\em International Conference on Machine Learning}, pp. 12888--12900. PMLR, 2022.

\bibitem{li2021align}
J.~Li, R.~Selvaraju, A.~Gotmare, S.~Joty, C.~Xiong, and S.~C.~H. Hoi.
\newblock Align before fuse: Vision and language representation learning with momentum distillation.
\newblock {\em Advances in neural information processing systems}, 34:9694--9705, 2021.

\bibitem{li2023does}
Y.~Li, J.~Wang, X.~Dai, L.~Wang, C.-C.~M. Yeh, Y.~Zheng, W.~Zhang, and K.-L. Ma.
\newblock How does attention work in vision transformers? a visual analytics attempt.
\newblock {\em IEEE Transactions on Visualization and Computer Graphics}, 2023.

\bibitem{coco}
T.-Y. Lin, M.~Maire, S.~Belongie, J.~Hays, P.~Perona, D.~Ramanan, P.~Dollár, and C.~L. Zitnick.
\newblock Microsoft coco: Common objects in context.
\newblock {\em arXiv preprint arXiv:1405.0312}, 2014.

\bibitem{ming2019protosteer}
Y.~Ming, P.~Xu, F.~Cheng, H.~Qu, and L.~Ren.
\newblock Protosteer: Steering deep sequence model with prototypes.
\newblock {\em IEEE transactions on visualization and computer graphics}, 26(1):238--248, 2019.

\bibitem{openai2023gpt4}
OpenAI.
\newblock Gpt-4 technical report, 2023.

\bibitem{papineni2002bleu}
K.~Papineni, S.~Roukos, T.~Ward, and W.-J. Zhu.
\newblock Bleu: a method for automatic evaluation of machine translation.
\newblock In {\em Proceedings of the 40th annual meeting of the Association for Computational Linguistics}, pp. 311--318, 2002.

\bibitem{park2019sanvis}
C.~Park, I.~Na, Y.~Jo, S.~Shin, J.~Yoo, B.~C. Kwon, J.~Zhao, H.~Noh, Y.~Lee, and J.~Choo.
\newblock Sanvis: Visual analytics for understanding self-attention networks.
\newblock In {\em 2019 IEEE VIS}, pp. 146--150. IEEE, 2019.

\bibitem{radford2021learning}
A.~Radford, J.~W. Kim, C.~Hallacy, A.~Ramesh, G.~Goh, S.~Agarwal, G.~Sastry, A.~Askell, P.~Mishkin, J.~Clark, et~al.
\newblock Learning transferable visual models from natural language supervision.
\newblock In {\em International conference on machine learning}, pp. 8748--8763. PMLR, 2021.

\bibitem{ren2016squares}
D.~Ren, S.~Amershi, B.~Lee, J.~Suh, and J.~D. Williams.
\newblock Squares: Supporting interactive performance analysis for multiclass classifiers.
\newblock {\em IEEE transactions on visualization and computer graphics}, 23(1):61--70, 2016.

\bibitem{ren2015faster}
S.~Ren, K.~He, R.~Girshick, and J.~Sun.
\newblock Faster r-cnn: Towards real-time object detection with region proposal networks.
\newblock {\em Advances in neural information processing systems}, 28, 2015.

\bibitem{ribeiro2016should}
M.~T. Ribeiro, S.~Singh, and C.~Guestrin.
\newblock " why should i trust you?" explaining the predictions of any classifier.
\newblock In {\em Proceedings of the 22nd ACM SIGKDD international conference on knowledge discovery and data mining}, pp. 1135--1144, 2016.

\bibitem{selvaraju2017grad}
R.~R. Selvaraju, M.~Cogswell, A.~Das, R.~Vedantam, D.~Parikh, and D.~Batra.
\newblock Grad-cam: Visual explanations from deep networks via gradient-based localization.
\newblock In {\em Proceedings of the IEEE international conference on computer vision}, pp. 618--626, 2017.

\bibitem{smilkov2016embedding}
D.~Smilkov, N.~Thorat, C.~Nicholson, E.~Reif, F.~B. Vi{\'e}gas, and M.~Wattenberg.
\newblock Embedding projector: Interactive visualization and interpretation of embeddings.
\newblock {\em stat}, 1050:16, 2016.

\bibitem{strickland2022andrew}
E.~Strickland.
\newblock Andrew ng, ai minimalist: The machine-learning pioneer says small is the new big.
\newblock {\em IEEE Spectrum}, 59(4):22--50, 2022.

\bibitem{strobelt2018s}
H.~Strobelt, S.~Gehrmann, M.~Behrisch, A.~Perer, H.~Pfister, and A.~M. Rush.
\newblock Seq2seq-vis: A visual debugging tool for sequence-to-sequence models.
\newblock {\em IEEE Trans. Vis. Comput. Graphics}, 25(1):353--363, 2018.

\bibitem{strobelt2021genni}
H.~Strobelt, J.~Kinley, R.~Krueger, J.~Beyer, H.~Pfister, and A.~M. Rush.
\newblock Genni: Human-ai collaboration for data-backed text generation.
\newblock {\em IEEE Transactions on Visualization and Computer Graphics}, 28(1):1106--1116, 2021.

\bibitem{touvron2023llama}
H.~Touvron, L.~Martin, K.~Stone, P.~Albert, A.~Almahairi, Y.~Babaei, N.~Bashlykov, S.~Batra, P.~Bhargava, S.~Bhosale, et~al.
\newblock Llama 2: Open foundation and fine-tuned chat models.
\newblock {\em arXiv preprint arXiv:2307.09288}, 2023.

\bibitem{vaswani2017attention}
A.~Vaswani, N.~Shazeer, N.~Parmar, J.~Uszkoreit, L.~Jones, A.~N. Gomez, L.~u. Kaiser, and I.~Polosukhin.
\newblock Attention is all you need.
\newblock In {\em Advances in Neural Information Processing Systems}, vol.~30, 2017.

\bibitem{vedaldi2008quick}
A.~Vedaldi and S.~Soatto.
\newblock Quick shift and kernel methods for mode seeking.
\newblock In {\em Computer Vision--ECCV 2008: 10th European Conference on Computer Vision, Marseille, France, October 12-18, 2008, Proceedings, Part IV 10}, pp. 705--718. Springer, 2008.

\bibitem{vedantam2015cider}
R.~Vedantam, C.~Lawrence~Zitnick, and D.~Parikh.
\newblock Cider: Consensus-based image description evaluation.
\newblock In {\em Proceedings of the IEEE conference on computer vision and pattern recognition}, pp. 4566--4575, 2015.

\bibitem{vig2019multiscale}
J.~Vig.
\newblock A multiscale visualization of attention in the transformer model.
\newblock In {\em Proceedings of the 57th Annual Meeting of the Association for Computational Linguistics: System Demonstrations}, pp. 37--42, 2019.

\bibitem{wang2018dqnviz}
J.~Wang, L.~Gou, H.-W. Shen, and H.~Yang.
\newblock Dqnviz: A visual analytics approach to understand deep q-networks.
\newblock {\em IEEE Trans. Vis. Comput. Graphics}, 25(1):288--298, 2018.

\bibitem{wang2023visual}
J.~Wang, S.~Liu, and W.~Zhang.
\newblock Visual analytics for machine learning: A data perspective survey.
\newblock {\em arXiv preprint arXiv:2307.07712}, 2023.

\bibitem{wang2022git}
J.~Wang, Z.~Yang, X.~Hu, L.~Li, K.~Lin, Z.~Gan, Z.~Liu, C.~Liu, and L.~Wang.
\newblock Git: A generative image-to-text transformer for vision and language.
\newblock {\em arXiv preprint arXiv:2205.14100}, 2022.

\bibitem{wangDodrioExploringTransformer2021}
Z.~J. Wang, R.~Turko, and D.~H. Chau.
\newblock Dodrio: {{Exploring Transformer Models}} with {{Interactive Visualization}}.
\newblock In {\em Proceedings of the 59th {{Annual Meeting}} of the {{Association}} for {{Computational Linguistics}} and the 11th {{International Joint Conference}} on {{Natural Language Processing}}: {{System Demonstrations}}}, pp. 132--141. {Association for Computational Linguistics}, {Online}, 2021.

\bibitem{yang2020interactive}
W.~Yang, X.~Wang, J.~Lu, W.~Dou, and S.~Liu.
\newblock Interactive steering of hierarchical clustering.
\newblock {\em IEEE Transactions on Visualization and Computer Graphics}, 27(10):3953--3967, 2020.

\bibitem{yang2022diagnosing}
W.~Yang, X.~Ye, X.~Zhang, L.~Xiao, J.~Xia, Z.~Wang, J.~Zhu, H.~Pfister, and S.~Liu.
\newblock Diagnosing ensemble few-shot classifiers.
\newblock {\em IEEE Transactions on Visualization and Computer Graphics}, 28(9):3292--3306, 2022.

\bibitem{yeh2023attentionviz}
C.~Yeh, Y.~Chen, A.~Wu, C.~Chen, F.~Vi{\'e}gas, and M.~Wattenberg.
\newblock Attentionviz: A global view of transformer attention.
\newblock {\em arXiv preprint arXiv:2305.03210}, 2023.

\bibitem{yuan2021survey}
J.~Yuan, C.~Chen, W.~Yang, M.~Liu, J.~Xia, and S.~Liu.
\newblock A survey of visual analytics techniques for machine learning.
\newblock {\em Computational Visual Media}, 7:3--36, 2021.

\bibitem{zhang2022sliceteller}
X.~Zhang, J.~P. Ono, H.~Song, L.~Gou, K.-L. Ma, and L.~Ren.
\newblock Sliceteller: A data slice-driven approach for machine learning model validation.
\newblock {\em IEEE Transactions on Visualization and Computer Graphics}, 29(1):842--852, 2022.

\bibitem{zhao2021human}
Z.~Zhao, P.~Xu, C.~Scheidegger, and L.~Ren.
\newblock Human-in-the-loop extraction of interpretable concepts in deep learning models.
\newblock {\em IEEE Transactions on Visualization and Computer Graphics}, 28(1):780--790, 2021.

\end{thebibliography}
